\documentclass[letterpaper, 10 pt, conference]{ieeeconf}  

\IEEEoverridecommandlockouts                              

\usepackage{graphics} 
\usepackage{epsfig} 
\usepackage{hyperref}
\usepackage{balance}
\usepackage{mathptmx} 
\usepackage{times} 
\usepackage{amsmath} 
\usepackage{amssymb}  
\usepackage{interval}
\usepackage{threeparttable}
\usepackage{booktabs}
\usepackage{adjustbox}
\usepackage{multicol}
\usepackage{lipsum}
\usepackage{graphicx}
\usepackage{booktabs}
\usepackage{caption}
\usepackage{pgfplots}
\pgfplotsset{compat=1.18}
\usepackage{pgfplotstable}
\usepackage{subcaption}
\usepackage{tikz}
\usepackage{array}
\usepackage{pgfplots}
\usepackage{bm}
\usepackage{float}
\usepackage{tabularx}
\usepackage{algorithm}
\usepackage{algpseudocode}
\usepackage{amsfonts}
\usepackage{multirow}
\usepackage{tikz}
\usepackage{xcolor}
\usepackage{pgfplots}

\definecolor{customred}{RGB}{255, 89, 94}  
\definecolor{customblue}{RGB}{25, 130, 196} 
\definecolor{custompurple}{RGB}{204, 71, 120}
\definecolor{customgreen}{RGB}{138, 201, 38}
\definecolor{customyellow}{RGB}{255, 202, 58}
\definecolor{custompurple2}{RGB}{106, 76, 147}

\title{\LARGE \bf
R900: Understanding the Cost-Effectiveness of Random Exploration from 900 Hours of Robotic Data Collection
}

\author{Shutong Jin$^{*}$, Axel Kaliff$^{*}$, Ruiyu Wang, Muhammad Zahid and Florian T. Pokorny
\thanks{$*$Equal contribution. The authors are with the School of Electrical Engineering and Computer Science, KTH Royal Institute of Technology
        {\tt\{shutong, akaliff, fpokorny\}@kth.se}. This work was partially supported by the Wallenberg AI, Autonomous Systems and Software Program (WASP) funded by the Knut and Alice Wallenberg Foundation. The computations were enabled by the supercomputing resource Berzelius provided by the National Supercomputer Centre at Linköping University and the Knut and Alice Wallenberg Foundation, Sweden.%
        }
}

\begin{document}

\maketitle
\thispagestyle{empty}
\pagestyle{empty}

\begin{abstract}
Data scarcity presents a key bottleneck for imitation learning in robotic manipulation. 
In this paper, we focus on random exploration data—actions and video sequences produced autonomously via motions to randomly sampled positions in the workspace—to investigate their potential as a cost-effective data source.
Our investigation follows two paradigms: 
(a) random actions, where we assess their feasibility for autonomously bootstrapping data collection policies, and (b) random exploration video frames, where we evaluate their effectiveness in pre-training parameter-dense networks with self-supervised learning objectives.
To minimize human supervision, we first develop a fully automated pipeline that handles episode labeling, termination, and resetting using cloud-based microservices for real-time monitoring.
Building on this, we present a large-scale study on the cost-effectiveness of real-world random exploration in a non-trivial two-layer stacking task, drawing on statistical results from 807 hours of random actions, 71 hours of random exploration video (1.28M frames), and 1,260 times of policy evaluation.
The dataset will be made publicly available and access to the robot environment with the automated pipeline is to be made accessible via cloud service for future research.
\end{abstract}

\section{Introduction}
The collection of large-scale real-world data for imitation learning remains a costly and labor-intensive process.
For example, RT-1~\cite{brohan2022rt} required 17 months of effort from a dedicated team, and BC-Z~\cite{jang2022bc} involved 7 operators working over 5 months. 
Meanwhile, the need for more training data is becoming an urgent demand with the continued scaling of models and the rising diversity of deployment environments.
This motivates us to focus on an often overlooked yet low-cost data source: random exploration data, where robots randomly interact with objects in the environment for future task deployment.
Such data can be generated autonomously in diverse settings 24/7, and may capture a wide range of object-environment interactions under real-world physics.
We therefore ask: 
\textit{To what extent can random exploration data serve as a \textbf{cost-effective} source for imitation learning?}

\begin{figure}[t]
    \centering
  \begin{subfigure}[h]{0.48\textwidth}
    \centering
    \includegraphics[width=\textwidth]{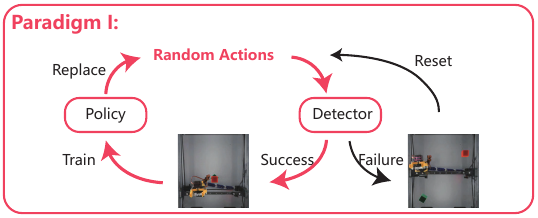}
    \vspace{-1.8 em}
    \caption{}
    \label{fig:firstA}
  \end{subfigure}
  \begin{subfigure}[h]{0.48\textwidth}
    \centering
    \includegraphics[width=\textwidth]{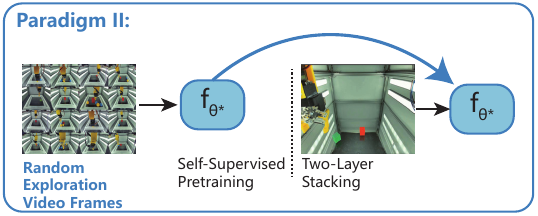} 
      \vspace{-1.8em}
    \caption{}
    \label{fig:firstB}
  \end{subfigure}
    \caption{\small(a) Episodes are generated autonomously via motions to randomly sampled positions in the workspace, with success or failure automatically labeled by a detector using a custom bottom camera beneath a transparent base plate. The successful episodes are used to train policies for autonomous data collection. (b) Random exploration video frames are used for self-supervised pre-training. The resulting pre-trained visual encoder $ f_{\theta^*} $ is then applied and evaluated in a behavior cloning policy for a two-layer stacking task.}
    \label{fig:introduction}
    \vspace{-1em}
\end{figure}

However, the unstructured nature and low task success rate of random exploration data create significant obstacles for its use in imitation learning.
The first obstacle is the absence of labeled information, whereas imitation learning typically relies on episodic data with observation-action pairs spanning a specific task.
This requires task-state labeling at every timestep, a process especially cumbersome in real-world settings.
The second obstacle is the heavy logistical burden of setting up large-scale real-world experimentation, where low task success rates of random exploration can require frequent (and even manual) scene resetting to restore misplaced objects.

To address these obstacles, we develop a fully automated pipeline for timestep-wise state labeling, episode termination, and scene resetting, supported by cloud-based microservices that process camera and joint signals for real-time monitoring.
Building on this, we first investigate the feasibility of autonomously bootstrapping data collection policies from purely random actions, without human supervision.
As shown in \textit{Fig.}~\ref{fig:firstA}, random actions are initially generated by motions to randomly sampled positions in the workspace, and episodes labeled as successful by the pipeline are then used to train a policy for autonomous data collection in later stages.
Our study is grounded in a fundamental two-layer stacking task, which provides a minimal form of sequential execution with two distinct skills—grasping and stacking.
With $1\,\text{cm}^3$ cubes randomly initialized on a $15 \times 15\,\text{cm}$ range, we analyze and visualize policy behaviors across various random action compositions, skill transferability, and cost-effectiveness relative to human demonstrations, supported by statistics from 807 hours of autonomous robot activity.

Alongside random actions generated via position sampling, we also study video frames collected during random exploration.
We hypothesize that, due to workspace similarity, certain visual features may still be learnable, such as object-environment interactions, visual variations, and overall scene layout.
As shown in \textit{Fig.}~\ref{fig:firstB}, we empirically investigate this hypothesis by curating a 71-hour planar pushing dataset of 1.28M frames, pre-training visual encoders with different self-supervised objectives, and evaluating their performance on a two-layer stacking task in the same environment.
Specifically, the self-supervised objectives include reconstruction loss MAE~\cite{he2022masked} to learn image patch relations, contrastive loss MoCo-v3~\cite{chen2021empirical} to separate static and dynamic features, and distillation loss DINO~\cite{caron2021emerging} to align local-global representations.
Our study first identifies the most effective self-supervised pre-training objective for random exploration video frames across architectures ranging from ViT-Base~\cite{dosovitskiy2020image} to ResNet18~\cite{he2016deep}, followed by an analysis of cost-effectiveness relative to other pre-training data sources~\cite {dasari2023unbiased,deng2009imagenet} and end-to-end strategies~\cite{mandlekar2021matters,chi2023diffusion}.
The main contributions of this work are:
\begin{itemize}
    \item We investigate the cost-effectiveness of real-world random exploration data for imitation learning through two paradigms: \MakeUppercase{\romannumeral 1}. \textbf{Random actions}, by assessing their feasibility for autonomously bootstrapping data collection policies. \MakeUppercase{\romannumeral 2}. \textbf{Random exploration video frames}, by evaluating their effectiveness in pre-training parameter-dense networks with self-supervised learning objectives.
    \item We develop a fully automated pipeline that handles episode labeling, termination, and resetting using cloud-based microservices for real-time monitoring.
    \item Based on statistics from 900 hours of real-world robot activities and 1,260 evaluations, our results suggest random actions as a cost-effective data source for imitation learning in the two-layer stacking scenario and provide strategies for managing the bootstrapping process. Within three policy updates, the time-efficiency gap with human keyboard demonstrations is reduced to 1.5 hours per 100 successful episodes, with an estimated cost of 0.53 USD. Additionally, we identify the contrastive objective with a ViT-Small architecture as an effective approach for utilizing random exploration video frames; however, its cost-effectiveness is reduced by the expensive training cost.

    \item We will release the dataset $R900$ containing 807 hours of random actions and 71 hours of random exploration video frames as an MIT-licensed open-source dataset. The robot environment, together with the automated pipeline, will be remotely accessible via cloud service to facilitate future research.
\end{itemize}

\section{Related Work}
\subsection{Robotic Manipulation Using Random Exploration Data}
The use of random exploration data has primarily focused on grasping~\cite{pinto2016supersizing, levine2018learning, detry2011learning, paolini2014data, morales2004using, pokorny2013grasp}, where a unique optimal label often does not exist due to the multiple ways an object can be grasped.
Prominent examples include \textit{Pinto et al.}~\cite{pinto2016supersizing}, who propose a staged data collection pipeline involving random grasps. Successful grasps detected by force sensors are used to train a Convolutional Neural Network (CNN)~\cite{krizhevsky2012imagenet} to predict grasp locations in the subsequent stages. 
\textit{Levine et al.}~\cite{levine2018learning} perform random grasps and use camera feedback to detect successful episodes, which are then used to train a success rate prediction network for visual servoing.
Other applications of random exploration data include training a pose regression CNN to initialize visual encoders~\cite{levine2016end} and developing inverse dynamics models~\cite{gao2024prime}.
In this study, we introduce real-time monitoring via cloud-based microservices~\cite{dragoni2017microservices} to extend random exploration data to imitation learning on sequential tasks (grasping, stacking) and analyze its cost-effectiveness as a data source.

\subsection{Self-Supervised Pre-training for Robotic Manipulation}
Self-supervised pre-training has been widely successful in various fields~\cite{radford2019language, radford2021learning} for its effectiveness in minimizing the need for labeled data.
Facing data scale challenges, the field of robotic manipulation has begun to explore self-supervised pre-training as a means of reducing reliance on large labeled datasets~\cite{xiao2022masked, ma2022vip, nair2022r3m, majumdar2023we, seo2023masked, radosavovic2023robot}.
Notable methods like MVP~\cite{xiao2022masked} and R3M~\cite{nair2022r3m} leverage reconstruction~\cite{he2022masked} and time-contrastive losses~\cite{sermanet2018time} for self-supervised pre-training on large-scale vision datasets~\cite{deng2009imagenet, grauman2022ego4d, goyal2017something}, demonstrating transferability to various robotic tasks.
Inspired by earlier successes, this paper explores a different scalable data source, random exploration video frames, to study whether pre-training gains from environmental similarity and to assess its cost-effectiveness.

\section{Problem Formulation}
This paper investigates the cost-effectiveness of random exploration data through two paradigms: Paradigm \MakeUppercase{\romannumeral 1} (random actions) and Paradigm \MakeUppercase{\romannumeral 2} (random exploration video frames), within the scope of imitation learning. 

\begin{figure}[h]
  \vspace{-0.5em}
  \centering
  \begin{subfigure}[h]{0.22\textwidth}
    \centering
    \includegraphics[width=\textwidth]{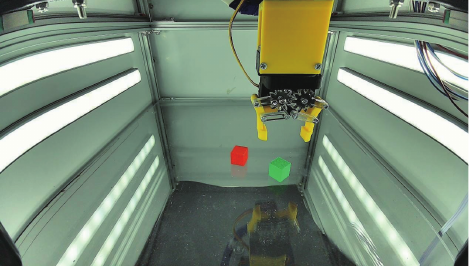}
    \vspace{-1.5em}
    \caption{}
    \label{fig:taskA}
  \end{subfigure}
  \hfill
  \begin{subfigure}[h]{0.25\textwidth}
    \centering
    \includegraphics[width=\textwidth]{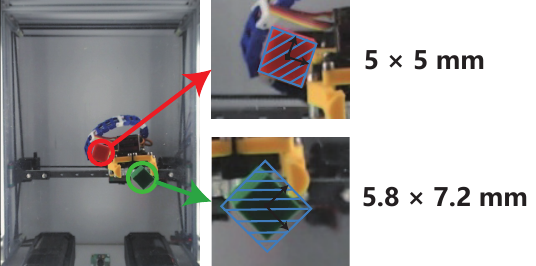} 
    \vspace{-1.5em}
    \caption{}
    \label{fig:taskB}
  \end{subfigure}
    \vspace{-0.5em}
    \caption{\small (a) Top-camera view $I^{t}$ of the workspace for policy training. 
    (b) Custom bottom-camera view $I^{b}$ of the workspace for pipeline automation, with hashed area showing policy fault tolerance range. The sizes \(5\,\text{mm} \times 5\,\text{mm}\) and \(5.8\,\text{mm} \times 7.2\,\text{mm}\) indicate the maximum allowable deviation from the ground-truth center for successful grasping and stacking, respectively.
    }
    \label{fig:task}
    \vspace{-1em}
\end{figure} 

\begin{figure*}[t] %
  \centering
  \begin{subfigure}[b]{0.34\textwidth}
    \centering
    \includegraphics[width=\textwidth]{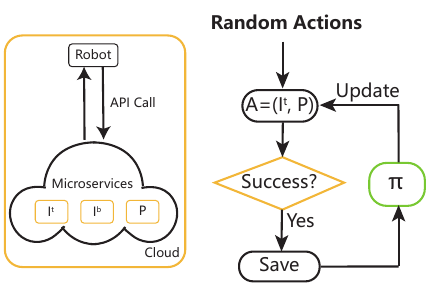}
    \caption{\small Paradigm \MakeUppercase{\romannumeral 1}}
    \label{fig:structureA}
  \end{subfigure}
  \hspace{0.02\textwidth}  
  \vrule width 0.5pt       
  \hspace{0.02\textwidth}
  \begin{subfigure}[b]{0.51\textwidth}
    \centering
    \includegraphics[width=\textwidth]{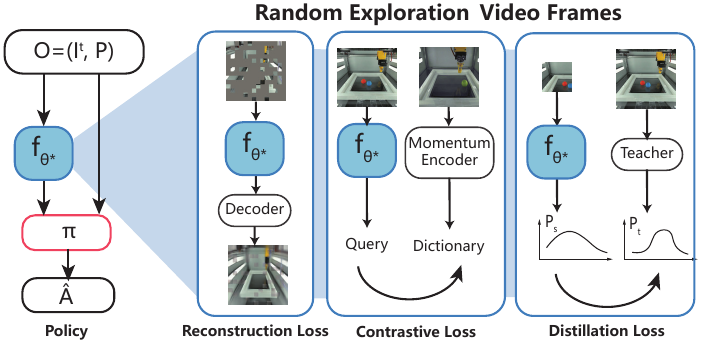} 
    \caption{\small Paradigm \MakeUppercase{\romannumeral 2}}
    \label{fig:structureB}
  \end{subfigure}
  \caption{\small(a) Illustration of Paradigm \MakeUppercase{\romannumeral 1}: In the first stage, the robot executes random actions $A$ and records the top-camera frames $ I^t $ and proprioception states $ P $ as an episode. Episode labeling, termination, and resetting are automated by the cloud-based microservices utilizing real-time sensory feedback ($I^t$, $I^b$, $P$). Stored successful episodes are used to train a policy for data collection in the second stage, which is performed iteratively. (b) Illustration of Paradigm \MakeUppercase{\romannumeral 2}: The behavior cloning policy $\pi$ takes the encoded top-camera frames $f_{\theta^*}(I^t)$ and proprioception states $P$ as inputs, and outputs actions $\hat{A}$. The visual encoder $f_{\theta^*}$ has three variants, each pre-trained on random exploration video frames using a different self-supervised objective: reconstruction, contrastive, or distillation loss. }
  \vspace{-1em}
\end{figure*}

\subsection{Paradigm \MakeUppercase{\romannumeral 1}: Random Actions}
\subsubsection{Definition}
Action sequences generated by moving to randomly sampled pick-up and drop-off positions within the workspace using position control APIs.

\subsubsection{Task}
\label{sec:task_description_paradigm1}
Grasp a randomly scattered $1\,\text{cm}^3$ green cube and stack it onto a $1\,\text{cm}^3$ red cube within a $15\;\text{cm}\;\times 15\;\text{cm}\times 3\;\text{cm}$ workspace (\textit{Fig.}~\ref{fig:taskA}).
We argue this task is non-trivial, given the wide range of possible initial cube positions and low tolerance for prediction faults shown in \textit{Fig.}~\ref{fig:taskB}. 
Due to the low likelihood of completing the full task using only random actions, the two-layer stacking task is further divided into two subtasks: 
Subtask {\romannumeral 1} (grasping) randomly samples a pick-up position to grasp the green cube, and Subtask {\romannumeral 2} (stacking) randomly samples a drop-off position to stack it onto the red cube.
Only 5\% and 2\% of random executions of Subtask~{\romannumeral 1} and Subtask~{\romannumeral 2}, respectively, succeed by chance.

\subsubsection{Dataset}
\label{sec:dataset_paradigm_1}
$R900_{\text{Action}}$ contains 807 hours of continuously recorded actions, comprising 700 successful and 11,561 failed episodes.
The episodes are organized by subtasks and their corresponding stages during data collection policy update, as summarized in \textit{Tab.}~\ref{tab:data_overview}.
Each recorded episode includes RGB frames from two cameras, proprioception states and actions, as detailed in \textit{Tab.}~\ref{tab:data_components}.

\renewcommand{\arraystretch}{1.2} 
\setlength{\tabcolsep}{8pt} 
\begin{table}[h]
\vspace{-1em}
\centering
\caption{\small Dataset overview for $R900_{\text{Action}}$.}
\label{tab:data_overview}
\begin{tabular}{llllll} 
\toprule
\textbf{Task} & \textbf{Stage} &{\textbf{Symbol}} & \textbf{Success} & \textbf{Total} \\ 
\midrule
\multirow{5}{*}{Subtask {\romannumeral 1}}  &Stage 1   & $S_1$  & 118& 2365  \\
&Stage 2 & $S_2$ & 210& 1750 \\
&{Stage 3}& $S_3^{\{1, 2\}}$ & 14& 100\\
&&$S_3^{\{1, 2\}E}$ &18& 100 \\
&&$S_3^{\{1\}}$ & 12& 100 \\
\midrule
\multirow{3}{*}{\parbox[c]{2cm}{Subtask {\romannumeral 1}\\\mbox{\hspace{1.7em}+}\\Subtask {\romannumeral 2}}}
&Stage 4   & $S_4$ &100 & 6663\\
&Stage 5 & $S_5$ & 81 & 571 \\
&Stage 6 & $S_6^{\{4, 5\}}$ & 44 & 203 \\
&& $S_6^{\{4, 5\}E}$ & 45 & 204 \\
&& $S_6^{\{4\}}$ & 58 & 205 \\
\bottomrule
\end{tabular}
\begin{tablenotes}
\footnotesize
\item \{$S_3^{\{1, 2\}}$, $S_3^{\{1, 2\}E}$, $S_3^{\{1\}}$\} and \{$S_6^{\{4, 5\}}$, $S_6^{\{4, 5\}E}$, $S_6^{\{4\}}$\} are variations differing in random action composition for Stage 3 and Stage 6, respectively, as described in \textit{Sec.}~\ref{sec:implementation_1}. 
\end{tablenotes}
\vspace{-1em}
\end{table}

\renewcommand{\arraystretch}{1.2} 
\setlength{\tabcolsep}{8pt} 
\begin{table}[h]
\centering
\caption{\small Symbol definitions for $R900_{\text{Action}}$.}
\label{tab:data_components}
\begin{tabular}{lll} 
\toprule
\textbf{Symbol} & \textbf{Description} & \textbf{Dimension} \\ 
\midrule
$n$ & Number of frames & 1 \\
$I^t$ & RGB top-camera frames & $1280 \times 720 \times 3 \times n$ \\
$I^b$ & RGB bottom-camera frames & $480 \times 640 \times 3 \times n$ \\
$P$ & Proprioception states & $5 \times n$ \\
$A$ & Actions & $5 \times n$ \\
$l$ & Success label & $1$ \\ 
\bottomrule
\end{tabular}
\vspace{-1em}
\end{table}

\subsection{Paradigm \MakeUppercase{\romannumeral 2}: Random Exploration Video Frames}
\subsubsection{Definition}
Video frames are recorded while the same robot arm performs unrelated tasks (planar pushing) within the same environment as the target task (two-layer stacking). 
Here, ``same environment" refers to a policy deployment setup where the camera maintains a fixed pose, and both the background within the camera's field of view and the illumination remain constant. 

\subsubsection{Task}
Extended two-layer stacking in \textit{Sec.}~\ref{sec:task_description_paradigm1} with a longer task sequence. The red cube needs to be grasped and placed in the center first, then grasp the green cube and stack it on top of the red cube.

\subsubsection{Dataset}
\label{sec:push_1k}
$R900_{\text{Video}}$ contains 71 hours (1.28M frames) of continuously recorded planar pushing videos. Collected in the same environment as shown in \textit{Fig.}~\ref{fig:task}, the dataset features a gripper randomly pushing various objects (balls and cubes). It provides top-camera views $I^t$ for self-supervised pre-training and bottom-camera views $I^b$ for object tracking during pushing, with dimensions aligned to \textit{Tab.}~\ref{tab:data_components}.

\section{Method}
In this section, we first clarify the behavior cloning policy and task notation shared by Paradigms \MakeUppercase{\romannumeral 1} and \MakeUppercase{\romannumeral 2}, and subsequently describe the two paradigms proposed in this paper.
\subsection{Preliminary} 
In imitation learning, a policy $\pi$ learns an action distribution from expert demonstration $(I^t,\;P,\;A)$, where $I^t$ denotes RGB images, $P$ denotes proprioception states, and $A$ represents the actions. The policy $\pi$ maps $I^t$ and $P$ to predicted actions $\hat{A}$, as $\hat{A} = \pi(f_{\theta}(I^t),\;P)$, where $f_{\theta}$ is a visual encoder. The policy is trained by minimizing $\|A - \hat{A}\|$.

\subsection{Paradigm \MakeUppercase{\romannumeral 1}: Random Actions}
\label{subsec:trial-and-error}
As shown in \textit{Fig.}~\ref{fig:structureA}, Paradigm \MakeUppercase{\romannumeral 1} explores the feasibility of using random actions for autonomously bootstrapping a data collection policy for two-layer stacking.
As outlined in \textit{Tab.}~\ref{tab:data_overview}, the bootstrapping process is divided into six stages $S_{i},\; i \in \{1, 2, 3, 4, 5, 6\}$. 
During $S_{1}$, a dataset $\mathcal{D}_{1}$ consisting 50 successful episodes under random actions is collected.  In subsequent stages ($i \geq 2$), both the dataset $\mathcal{D}_{i}$ and the policy $\pi_{i}$ are updated according to the four steps described below:
\begin{enumerate}
    \item \textbf{Policy Update:} The policy $ \pi_i $ is trained using the dataset from the previous stage, $ \mathcal{D}_{i-1} $.

    \item \textbf{Data Collection:} The trained data collection policy $ \pi_i $ is used to autonomously generate actions to collect data. Successful episodes, identified by an automatic error detection toolkit, are collected and stored as $ \mathcal{D}_i $.

    \item \textbf{Data Processing:} A data distribution balancing operation is designed to even out the distribution of successful episodes within the dataset. Specifically, 50 episodes from $ \mathcal{D}_{i-1} $ are selected by maximizing the average pairwise $ L_1 $ distance between the initial positions $p$ of the green cube in $ \mathcal{D}_{i-1} $ and $ \mathcal{D}_i $:
    \begin{equation}
        \mathcal{D}_{i-1}' = \arg\max_{\substack{\mathcal{D}_{i-1}' \subseteq \mathcal{D}_{i-1} \\ |\mathcal{D}_{i-1}'| = 50}} \sum_{p_i \in \mathcal{D}_{i-1}'} \sum_{p_j \in \mathcal{D}_i} \|p_i - p_j\|_1,
        \label{eq:even}
    \end{equation}
    
    \item \textbf{Data Composition:} The dataset $ \mathcal{D}_i $ is merged with selected episodes from prior stages to increase data size:
    \begin{equation}
        \mathcal{D}_{i} =  \mathcal{D}_{i-1}' \cup \mathcal{D}_i.
    \end{equation}
\end{enumerate}

$S_1$ and $S_4$ use random actions for Subtasks {\romannumeral 1} and Subtask
 {\romannumeral 2}, respectively.
In $S_1$, a pick-up position $(x, y)$ is randomly sampled within the workspace. The gripper is lifted to $z=1$, moved to $(x, y)$, lowered to $z=0$, and then closed.
In $S_4$, the best-performing policy in $S_3$ performs Subtask {\romannumeral 1} (grasping), while Subtask {\romannumeral 2} (stacking) moves the green cube to a randomly sampled drop-off position $(x, y, z)$.
Throughout all stages, data are managed by a cloud-based platform that handles episode labeling, termination, and resetting.
The sensory inputs, comprising top-camera view $I^t$, bottom-camera view $I^b$, and proprioceptive states $P$, are streamed to the cloud for functions such as episode management, object tracking, and task state verification against pre-defined success patterns (e.g., object tracked in the closed gripper).
These functionalities are further split into independent microservices, which communicate over the network using asynchronous processing to achieve real-time monitoring.

\subsection{Paradigm \MakeUppercase{\romannumeral 2}: Random Exploration Video Frames}
As shown in \textit{Fig.}~\ref{fig:structureB}, camera frames $I^{t}$ from the dataset detailed in \textit{Sec.}~\ref{sec:push_1k} are used to pre-train a visual encoder $f_\theta$. A self-supervised objective function $\mathcal{L}_{SSL}$ is employed during the visual pre-training:

{
\small
\begin{equation}
    \theta^* = \arg \min_\theta \mathcal{L}_{SSL}(f_\theta(I^{t})),
\end{equation}
}
Three prevalent self-supervised objective functions are studied, namely $\mathcal{L}_{SSL}\in\{\mathcal{L}_{MAE},\, \mathcal{L}_{MoCo},\, \mathcal{L}_{DINO}\}$.

\textbf{Reconstruction Loss.} \underline{M}asked \underline{A}uto\underline{e}ncoders (MAE)~\cite{he2022masked} mask random patches of the input frame $I^{t}$ to create $I^{t}_{\text{masked}}$ and reconstructs the missing pixels using an asymmetric encoder-decoder architecture. The loss is defined as:
    {
    \small
     \begin{equation}
            \mathcal{L}_{\text{MAE}} = \frac{1}{|M|} \sum_{i \in M} \left\| \left[ \mathcal{F}_{\text{dec}}\left( f_{\theta^*}\left(I^{t}_{\text{masked}} \right) \right) \right]_i - I^{t}_i \right\|^2,
    \end{equation}    
    }
    where $M = \{ i \mid M_i = 1 \}$ denotes the set of indices corresponding to masked patches, with $|M|$ representing the number of masked patches; $\mathcal{F}_\text{dec}$ is a lightweight decoder.

\textbf{Contrastive Loss}. \underline{Mo}mentum \underline{Co}ntrast (MoCo v3)~\cite{chen2021empirical} pulls similar examples closer and pushes dissimilar ones apart using a momentum encoder and a dynamic dictionary. The loss based on infoNCE~\cite{oord2018representation} is defined as:
    {\small
    \begin{equation}
        \mathcal{L}_{\text{MoCo}} = -\log \frac{\exp(f_{\theta^*}(I^{t}) \cdot k^+ / \tau)}{\exp(f_{\theta^*}(I^{t}) \cdot k^+ / \tau) + \sum\limits_{k^-}{\exp(f_{\theta^*}(I^{t}) \cdot k^- / \tau)} },
    \end{equation}
    }
    where  $k^+$ corresponds to the positive sample (same frame as the query), $k^-$ are negative samples from unrelated frames, and $\tau$ is a temperature hyperparameter~\cite{wu2018unsupervised}.
    
     \textbf{Distillation Loss}. Self-\underline{di}stillation with \underline{no} labels (DINO)~\cite{caron2021emerging} trains a student network to match the outputs of a teacher network. For input frames $I^{t}$, the student processes global and local croppings of $I^{t}_s$, while the teacher processes only global croppings of $I^{t}_t$. The loss is defined as:
    { \small 
    \begin{equation} \mathcal{L}_{\text{DINO}} = - P_t(\mathcal{T}(I^{t}_t)) \cdot \log P_s(f_{\theta^*}(I^{t}_s)), \end{equation}
    }
    where $\mathcal{T}$ denotes the teacher network, and $P_s$ and $P_t$ represent the $K$-dimensional distributions produced by the student and teacher networks, respectively.

\section{Experiments}
\label{sec:experiment}

\subsection{Experimental Setup}
\subsubsection{Paradigm \MakeUppercase{\romannumeral 1}}
\label{sec:implementation_1}

\paragraph{Implementation Details}
Adding to \textit{Sec.}~\ref{subsec:trial-and-error}, $S_3$ and $S_6$ each include three variations based on different data compositions of random actions.  
For $S_3$:  
(a) $S_3^{\{1,2\}}$: $\mathcal{D}_3^{\{1,2\}}$ is collected using $\pi_3^{\{1,2\}}$, trained on 50 successful episodes from $\mathcal{D}_1$ and 50 successful episodes $\mathcal{D}_2$;  
(b) $S_3^{\{1,2\}E}$: $\mathcal{D}_3^{\{1,2\}E}$ is collected using $\pi_3^{\{1,2\}E}$, trained on the same combination of data as $S_3^{\{1,2\}}$ but with an enforced even distribution from $\mathcal{D}_1$ using \textit{Eq.}~\ref{eq:even}.  
(c) $ S_3^{\{1\}} $: $\mathcal{D}_3^{\{1\}}$ collected using $\pi_3^{\{1\}}$, trained on 100 episodes from $\mathcal{D}_1$.
For $S_6$, the variants $S_6^{\{4,5\}}$, $S_6^{\{4,5\}E}$, and $S_6^{\{4\}}$ follow the same methodology as $S_3$ but use data sourced from $\mathcal{D}_4$ and $\mathcal{D}_5$. 
Policy $\pi$ is behavior cloning with Gaussian mixtures, using hyperparameters in prior studies~\cite{dasari2023unbiased, mandlekar2021matters, rahmatizadeh2018virtual}: five kernels, a learning rate of 1e-4, L2 weight decay of 1e-4 and 50,000 iterations. 
We choose this policy for its widespread adoption and low data needs, which align with the low likelihood of successful episodes from pure random actions in the initial stages.

\paragraph{Evaluation Protocol}
Success rates are computed from 100 trials per policy and normalized to the range $[0,1]$, with 1,000 total evaluations.
We report results for Subtask~{\romannumeral 1} and for the complete task (Subtask {\romannumeral 1} + Subtask {\romannumeral 2}) separately.

\begin{figure*}[t]
    \centering
    \includegraphics[width=0.7\linewidth]{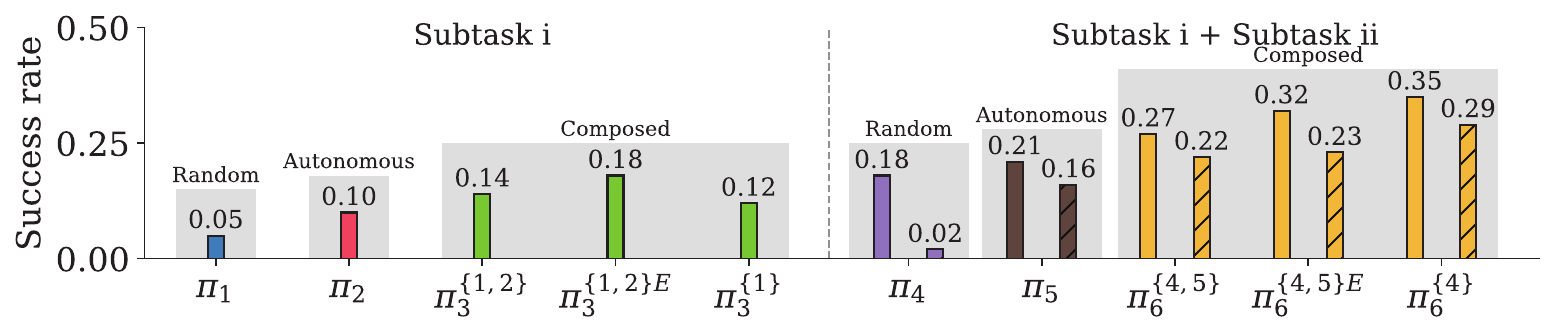}
    \vspace{-1em}
    \caption{\small Success rates for the data collection policies. Solid bars indicate Subtask {\romannumeral 1} results, and hatched bars indicate complete-task results.
    }
    \label{fig:success_rate}
    \vspace{-1em}
\end{figure*}

\subsubsection{Paradigm \MakeUppercase{\romannumeral 2}}
\paragraph{Implementation Details}
\label{sec:implementation_details_paradigm2}
During pre-training, three self-supervised objectives are applied to Vision Transformer~\cite{dosovitskiy2020image} (ViT-Base, ViT-Small) and ResNet~\cite{he2016deep} (ResNet50, ResNet18), using 1.28M, 700K, 700K, and 300K random exploration video frames, respectively. 
Most hyperparameters follow the original papers~\cite{he2022masked, chen2021empirical, caron2021emerging}, with only the learning rate adjusted (1e-5) to account for differences in image diversity between random exploration video frames and ImageNet~\cite{deng2009imagenet}. 
Following prior work on self-supervised pre-training in robotic manipulation~\cite{dasari2023unbiased, burns2023makes, parisi2022unsurprising}, we adopt behavior cloning with Gaussian mixtures as the policy used in \textit{Tab.}~\ref{table:architecture}, \ref{table:size}, and \ref{table:cost_effectiveness_paradigm2}, with hyperparameters identical to those listed in \textit{Sec.}~\ref{sec:implementation_1}. In \textit{Tab.}~\ref{table:cost_effectiveness_paradigm2}, diffusion policy~\cite{chi2023diffusion} is implemented using the standard hybrid convolutional neural network with 2-step observations, a 16-step prediction horizon, and 8-step action executions, following the hyperparameters from the original paper.
Each training uses 200 expert human demonstration episodes.

\paragraph{Evaluation Protocol} We evaluate policy performance using four metrics: Prediction Error (\({PE}\)), the mean absolute difference between predicted actions and ground truth; \(S_{\text{grasping}}\), the success rate of gripper grasping the red cube; \(S_{\text{one}}\), the success rate of picking and placing the red cube at the center; and \(S_{\text{two}}\), the success rate of stacking the green cube on top of the red cube. Each policy is evaluated 20 times with varied initial cube positions.

\subsection{Results of Paradigm \MakeUppercase{\romannumeral 1}}
\label{sec:random_commands}
As outlined in \textit{Tab.}~\ref{tab:data_overview}, the success rates of the corresponding policies are shown in \textit{Fig.}~\ref{fig:success_rate}.
\subsubsection{\textbf{What trends emerge in data collected by different policies?}}
\label{sec:trends_emerged}
\begin{figure}[h]
\centering
  \begin{subfigure}[h]{0.12\textwidth}
    \centering
    \includegraphics[width=\textwidth]{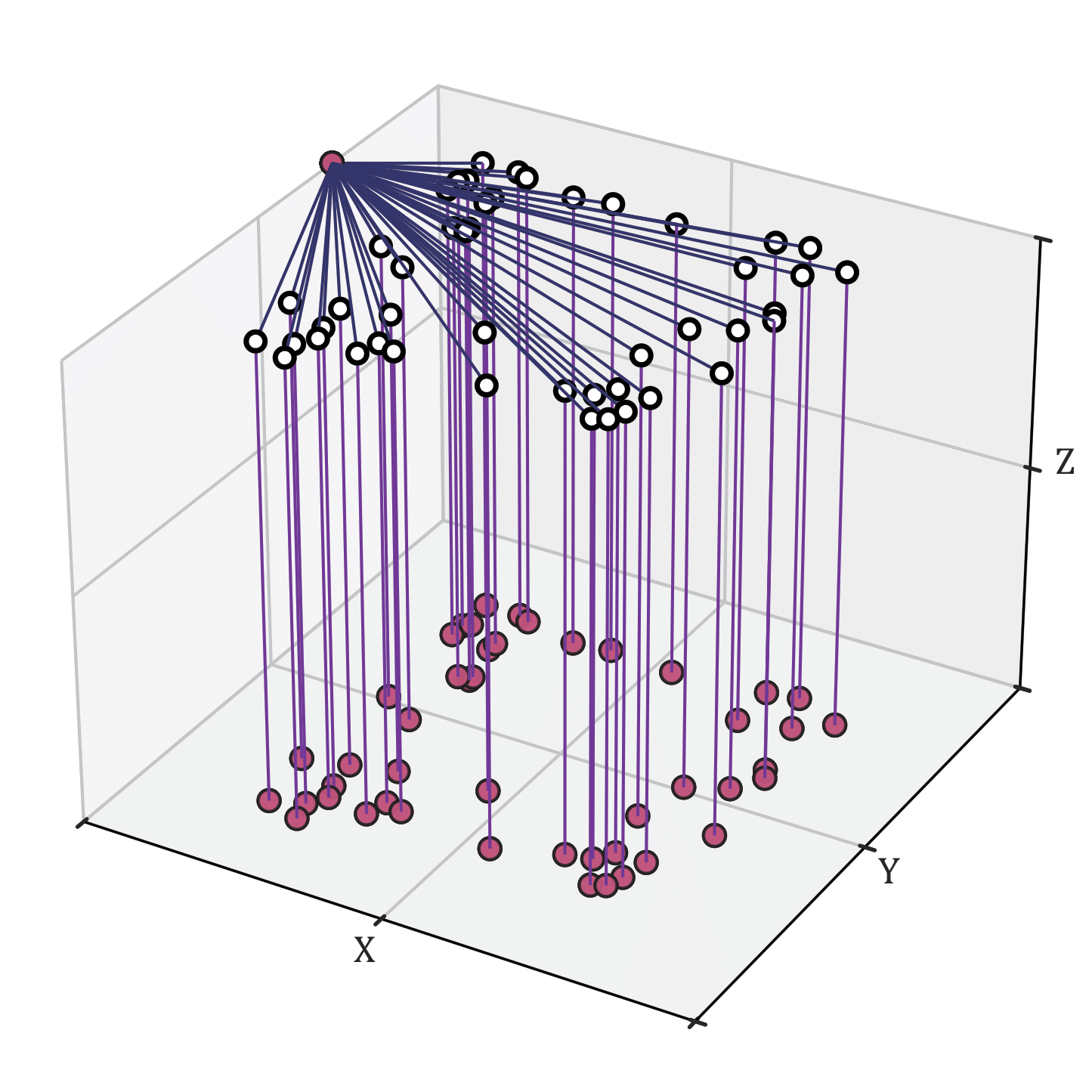}
    \subcaption*{$L_{1}^{\text{avg}}=0.0378$}
    \includegraphics[width=\textwidth]{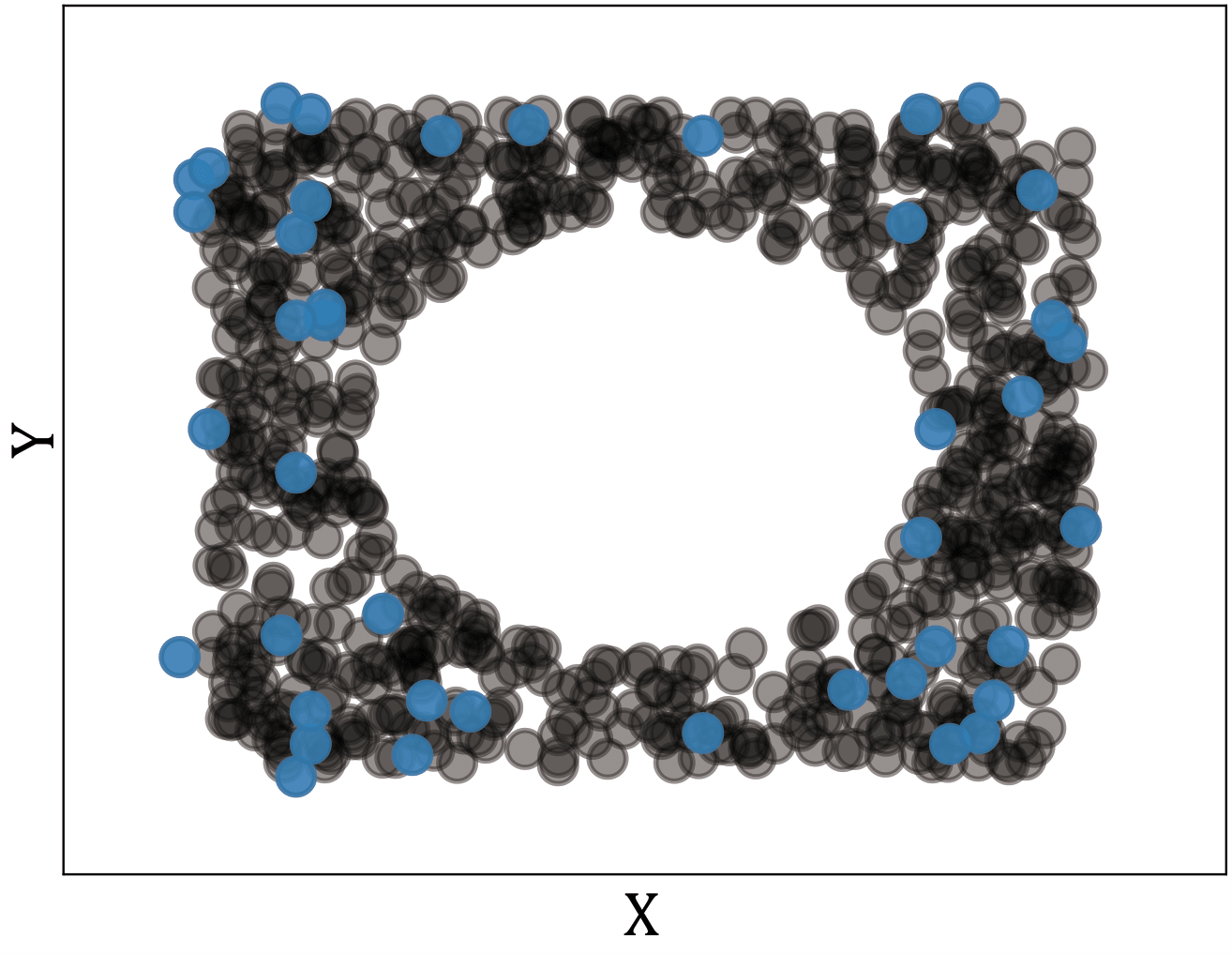}
    \includegraphics[width=\textwidth]{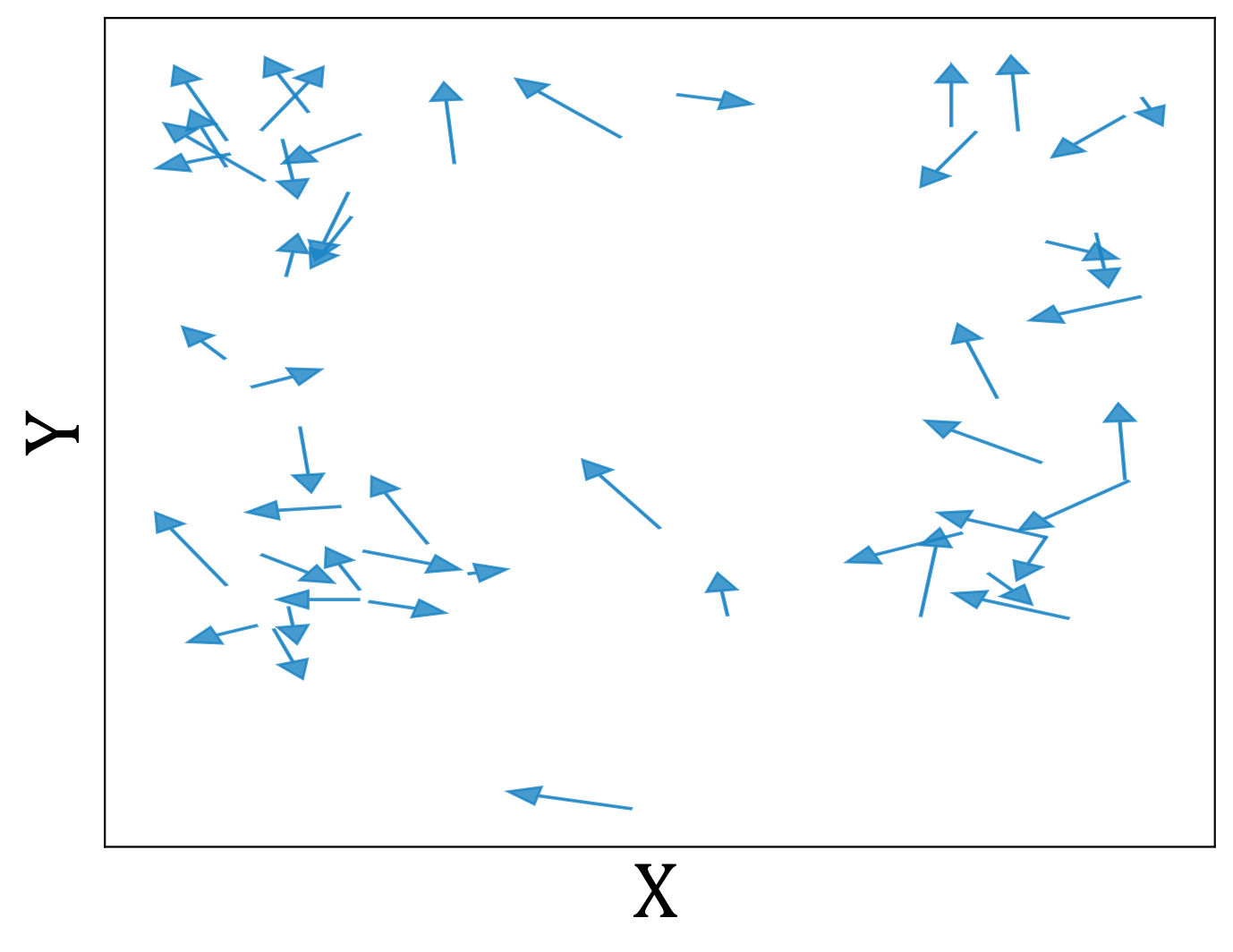}
    \caption{\small $\pi_1$}
    \label{fig:data_trendA}
  \end{subfigure}
  \begin{subfigure}[h]{0.12\textwidth}
    \centering
    \includegraphics[width=\textwidth]{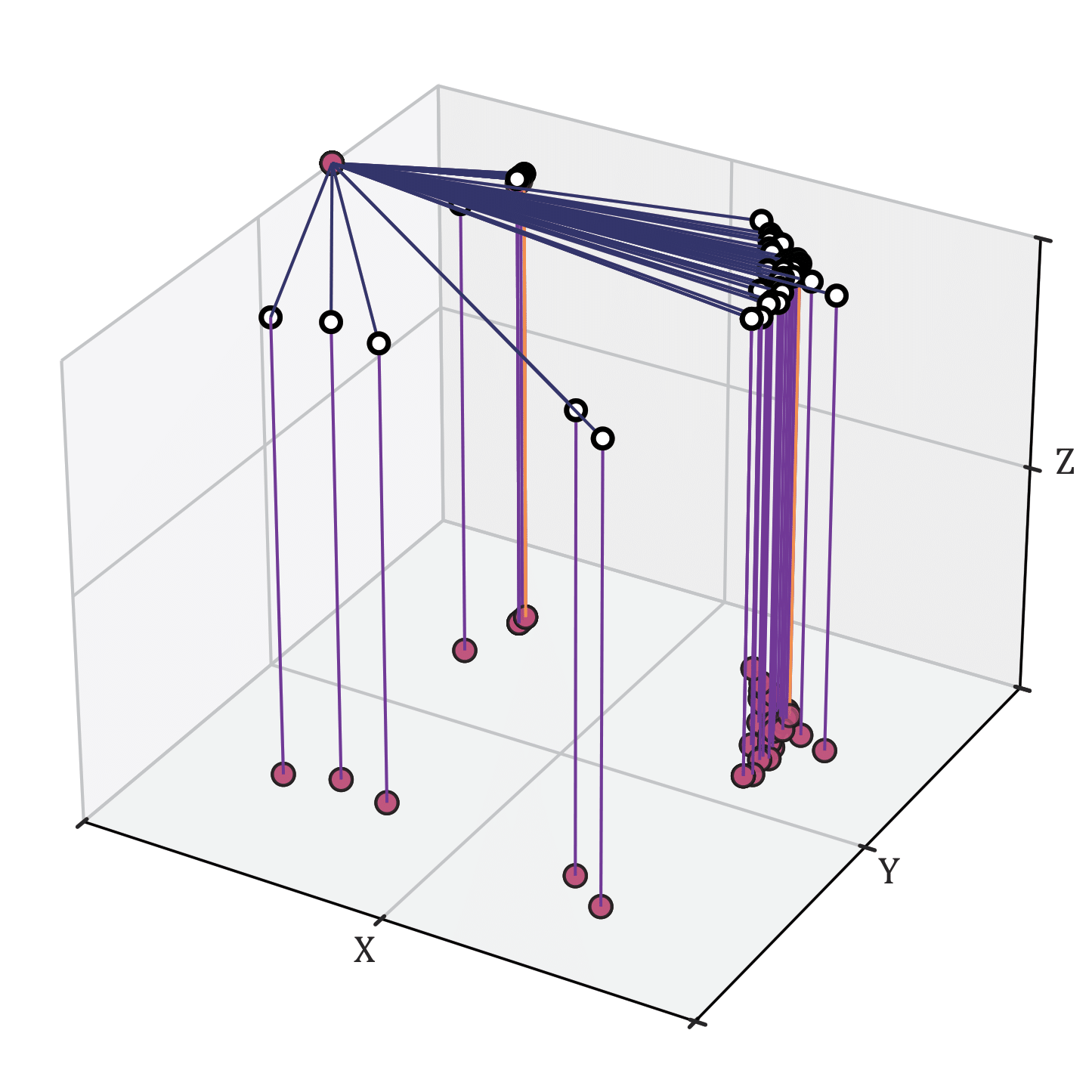}
    \subcaption*{$L_{1}^{\text{avg}}=0.0296$}
    \includegraphics[width=\textwidth]{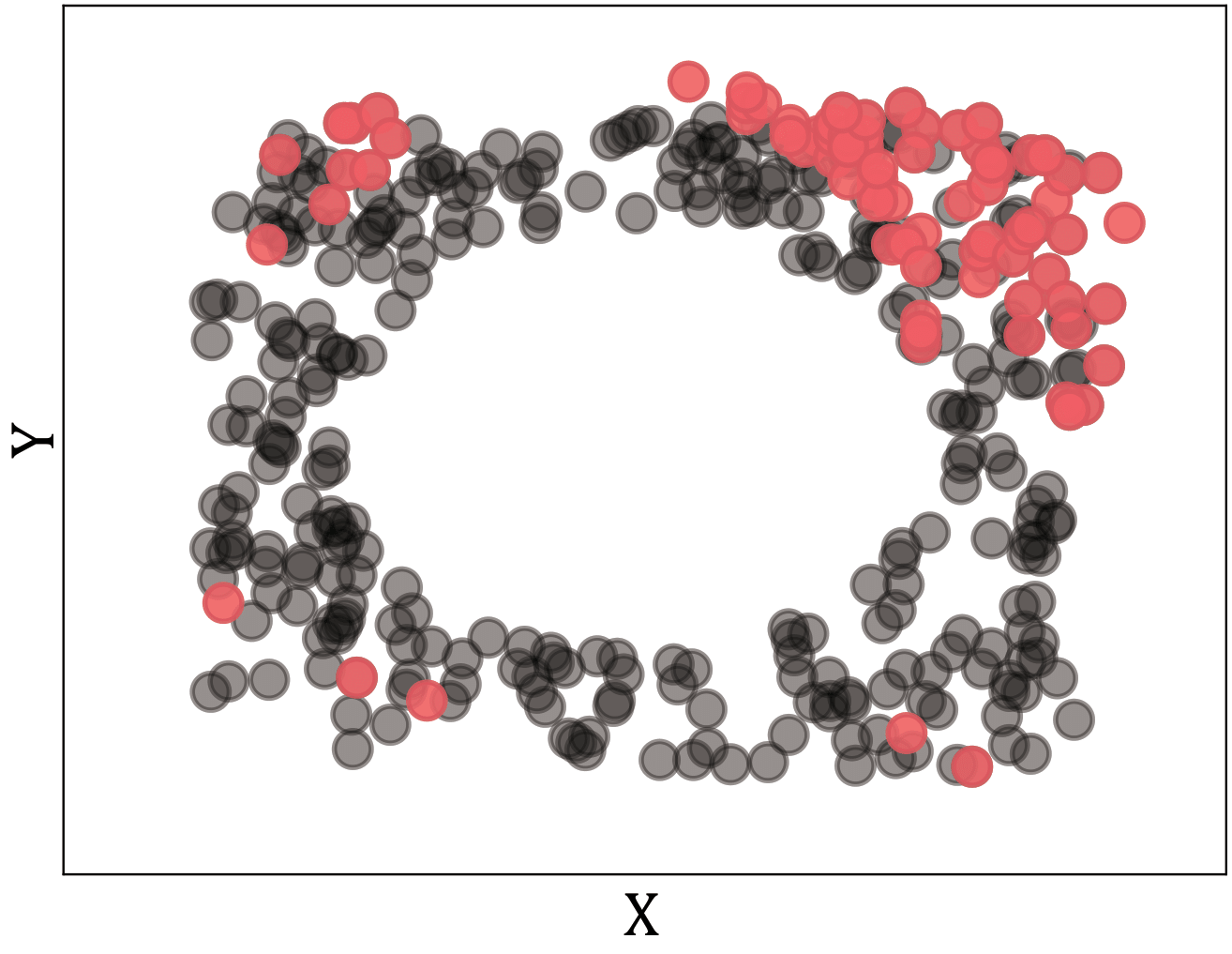} 
    \includegraphics[width=\textwidth]{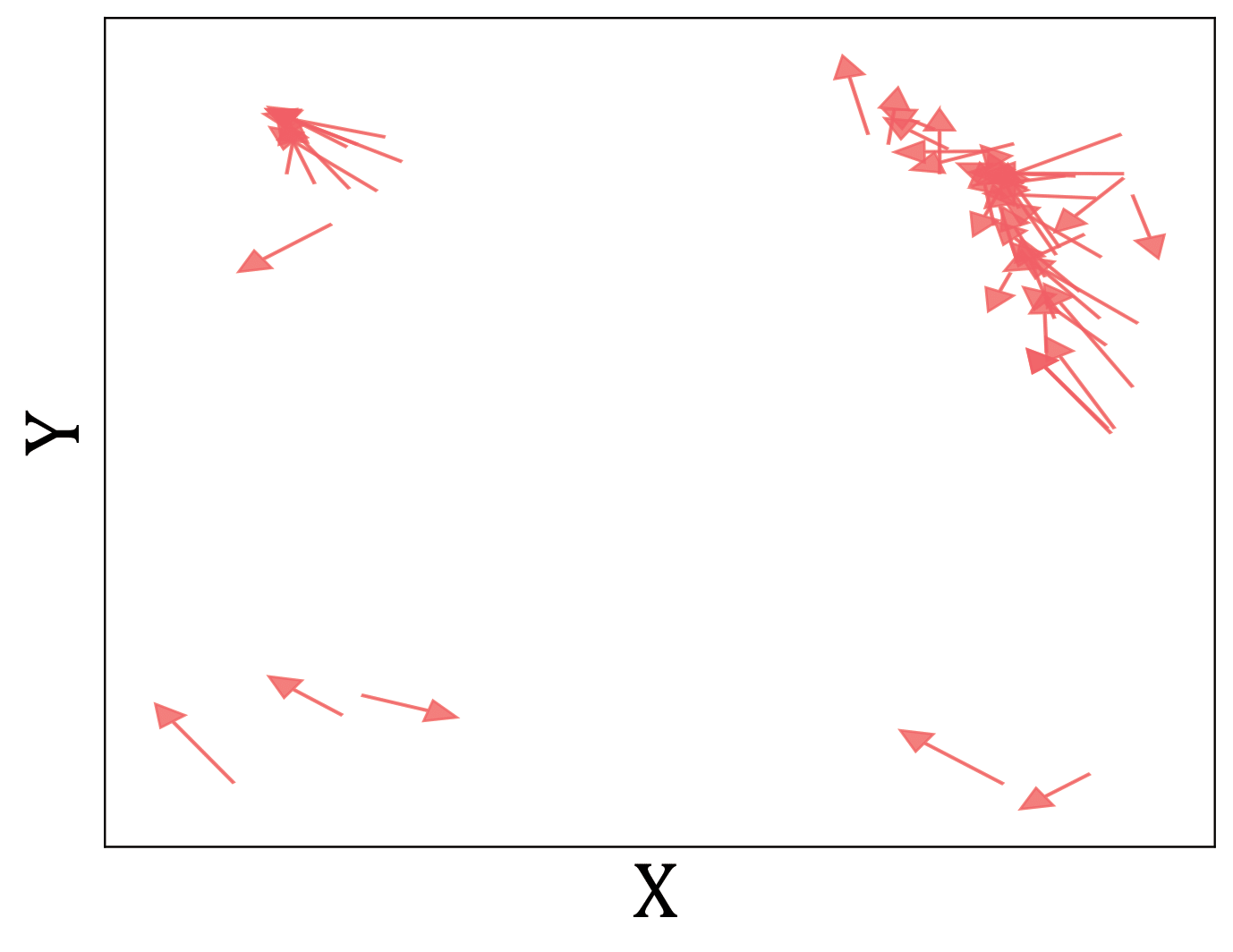}
    \caption{\small$\pi_2$}
    \label{fig:data_trendB}
  \end{subfigure}
  \begin{subfigure}[h]{0.12\textwidth}
    \centering
    \includegraphics[width=\textwidth]{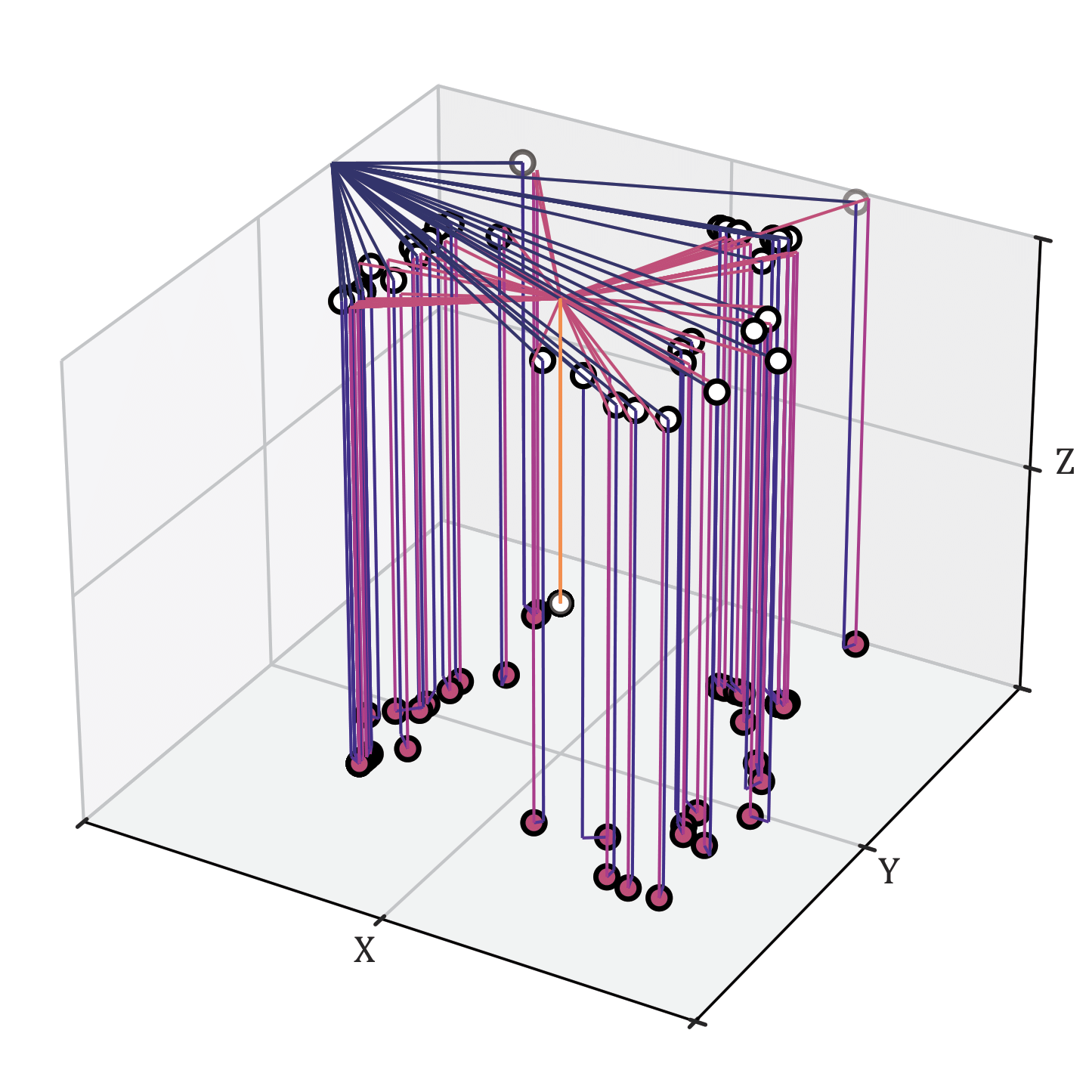}
    \subcaption*{$L_{1}^{\text{avg}}=0.0343$}
    \includegraphics[width=\textwidth]{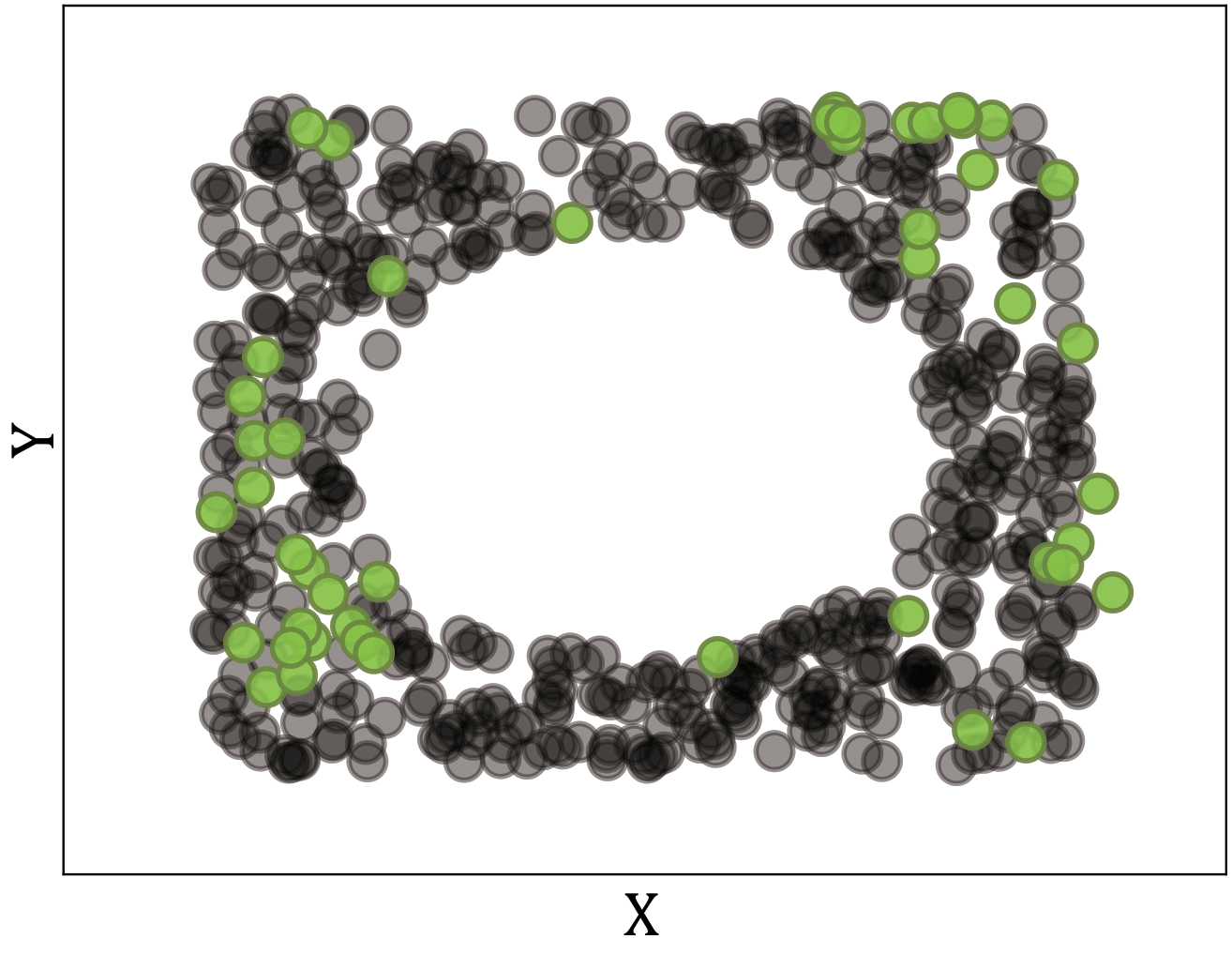} 
    \includegraphics[width=\textwidth]{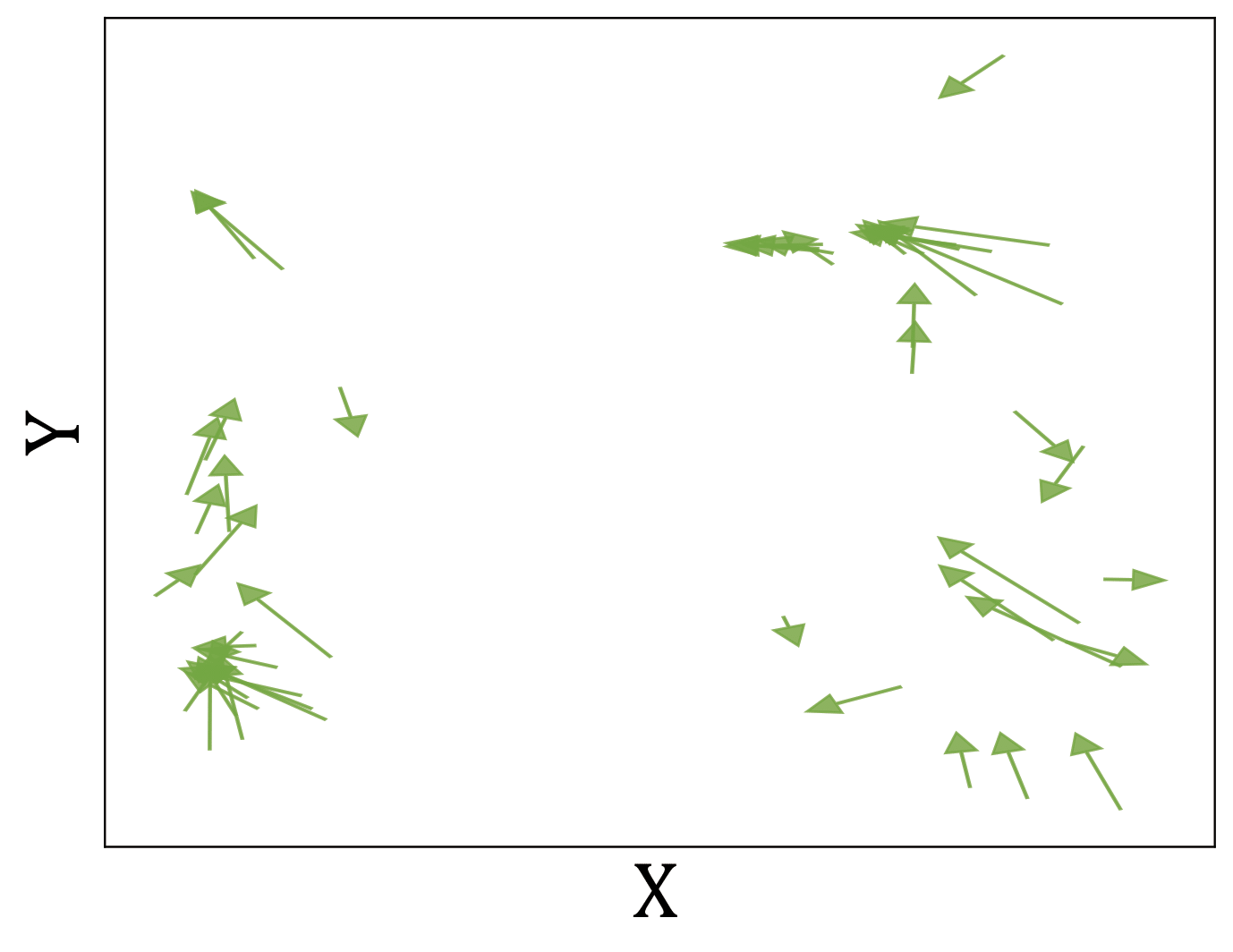}
    \caption{\small$\pi_3^{\{1, 2\}E}$ }
    \label{fig:data_trendC}
  \end{subfigure}
  \begin{tikzpicture}
        \draw[black, thick] (-4,-1) circle (3pt);
        \node[right] at (-3.9,-1) {Gripper Open};
        \draw[black, thick] (-1,-1) circle (3.5pt);
        \fill[custompurple, thick] (-1,-1) circle (3pt);
        \node[right] at (-0.9,-1) {Gripper Close};
        \fill[customblue] (-4,-1.5) circle (3pt);
        \node[right] at (-3.9,-1.5) {Stage 1};
        \fill[customred] (-2,-1.5) circle (3pt);
        \node[right] at (-1.9,-1.5) {Stage 2};
        \fill[customgreen] (0,-1.5) circle (3pt);
        \node[right] at (0.1,-1.5) {Stage 3};
        \fill[gray] (2,-1.5) circle (3pt);
        \node[right] at (2.1,-1.5) {Failed Episodes};
        \draw[<-, thick, customblue] (-4,-2) -- (-3.5,-2);
        \node[right] at (-3.4,-2) {Stage 1};
        \draw[<-, thick, customred] (-1,-2) -- (-0.5,-2);
        \node[right] at (-0.4,-2) {Stage 2};
        \draw[<-, thick, customgreen] (2,-2) -- (2.5,-2);
        \node[right] at (2.6,-2) {Stage 3};
    \end{tikzpicture}
    \caption{\small Data collected by $\pi_1$, $\pi_2$, and $\pi_3^{\{1, 2\}E}$. The first row shows trajectories of successful episodes with gripper open/close indicators. The second row shows the average pairwise $L_1$ distance and a top-down ($xy$-plane) view of the green cube’s initial position in all episodes (both successful and failed). The third row shows the difference between the predicted and ground truth green cube position in the successful episodes, with arrows indicating the direction from the ground truth to the predicted position. }
    \label{fig:data_trend}
\end{figure}

We first examine data collected by $\pi_1$ (random policy), $\pi_2$ (autonomous policy), and $\pi_3^{\{1,2\}E}$ (best-performing composed policy), evaluating robot activity and prediction discrepancy, as illustrated in \textit{Fig.}~\ref{fig:data_trend}.
Distinct trends emerge across the collected data: 
\begin{itemize}
    \item Data collected by $\pi_1$ has \textit{the most even distribution}, primarily due to the large volume of collected data. Note that during $S_1$, no episodes occurred in the center due to a 3 cm limit set around the red cube, preventing accidental disturbances from random actions.
    \item Data collected by $\pi_2$ shows \textit{a strong distributional bias}, with most successful episodes occurring when the green cube's initial position is in the top-right corner. We hypothesize that this results from a biased relative position between the predicted and ground truth green cube positions. As shown in the third row of \textit{Fig.}~\ref{fig:data_trendB}, predictions are often top-left of the ground truth position, potentially leading to higher success rates in certain regions. This bias is less apparent in data collected by the random policy $ \pi_1 $ and the composed policy $ \pi_3^{\{1, 2\}E} $.
    \item Data collected by $\pi_3$ achieves a more balanced distribution than $ \mathcal{D}_2 $ and $\pi_3^{\{1, 2\}E}$ attains \textit{the highest success rate} among all data collection policies in Subtask {\romannumeral 1}. This improvement may result from composing $\mathcal{D}_1$ and $\mathcal{D}_2$ with enforced even distribution, which increases the amount of training episodes while preserving distributional balance.
\end{itemize}

\subsubsection{\textbf{What role do random actions play in data composition?}}
\label{sec:lnn}
\begin{figure}[h]
\vspace{-2em}
\centering
  \begin{subfigure}[h]{0.12\textwidth}
    \centering
    \includegraphics[width=\textwidth]{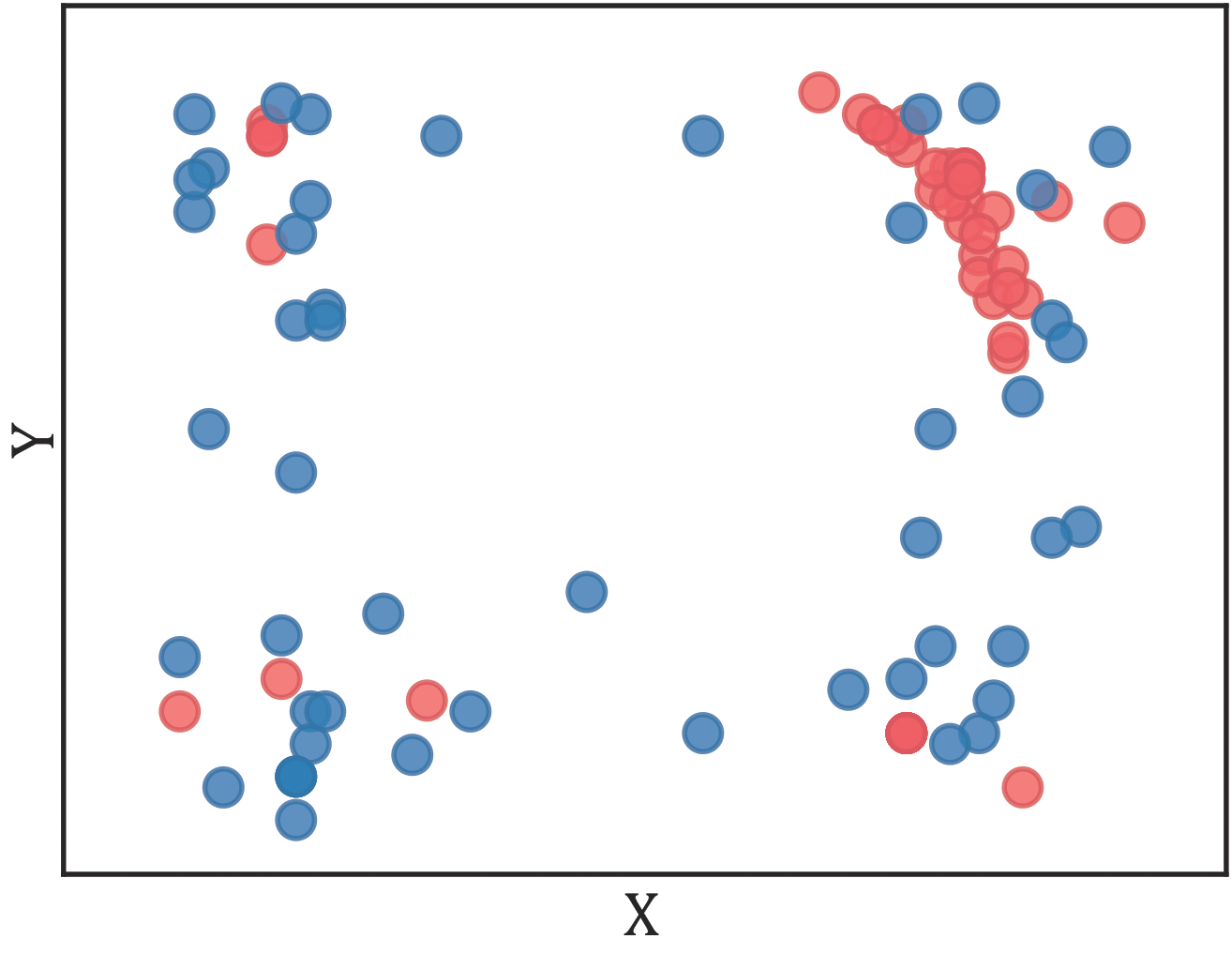}
    \caption{\small$\pi_3^{\{1, 2\}}$}
    \label{fig:data_compositionA}
  \end{subfigure}
  \begin{subfigure}[h]{0.12\textwidth}
    \centering
    \includegraphics[width=\textwidth]{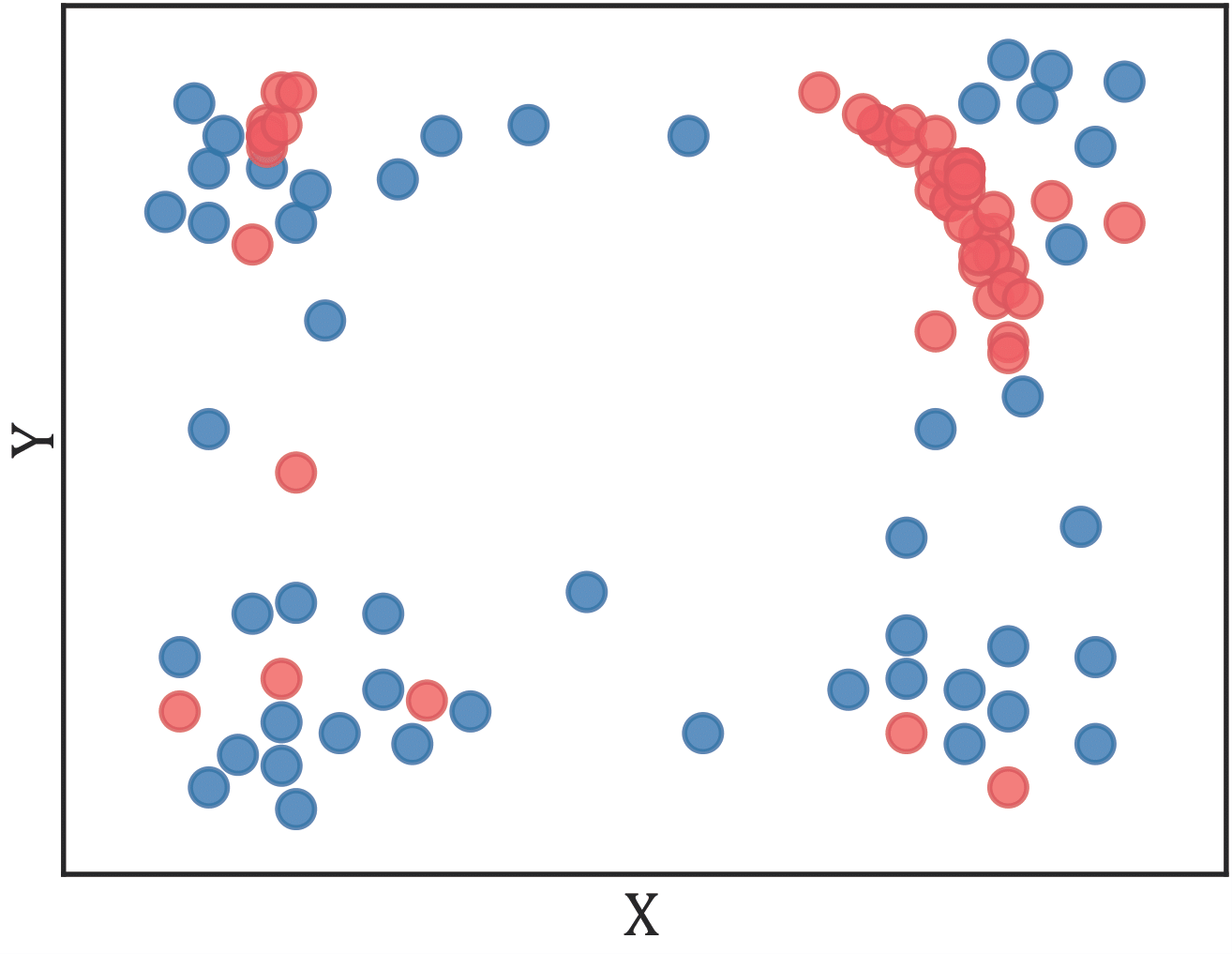} 
    \caption{\small$\pi_3^{\{1, 2\}E}$}
    \label{fig:data_compositionB}
  \end{subfigure}
  \begin{subfigure}[h]{0.12\textwidth}
    \centering
    \includegraphics[width=\textwidth]{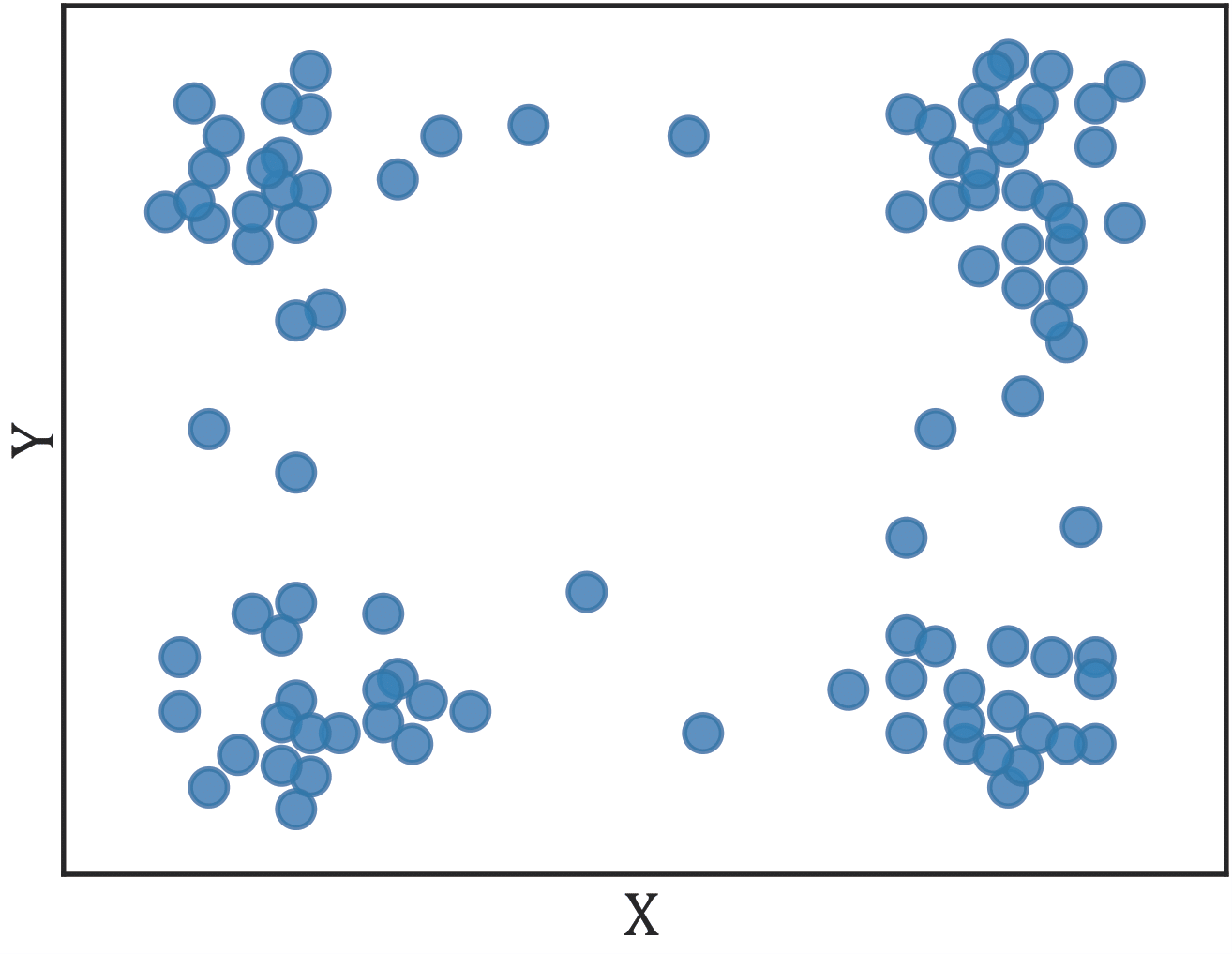} 
    \caption{\small$\pi_3^{\{1\}}$}
    \label{fig:data_compositionC}
  \end{subfigure}
  \begin{tikzpicture}
    \fill[customblue] (-9,-1) circle (3pt);
        \node[right] at (-8.9,-1) {Stage 1};
        \fill[customred] (-7,-1) circle (3pt);
        \node[right] at (-6.9,-1) {Stage 2};
    \end{tikzpicture}
    \vspace{-1em}
    \caption{\small Top-down ($xy$-plane) view of the green cube’s positions in the composed data used for training $\pi_3^{\{1, 2\}}$, $\pi_3^{\{1, 2\}E}$, and $\pi_3^{\{1\}}$.}
    \label{fig:data_composition}
\end{figure}

As detailed in \textit{Sec.}~\ref{sec:implementation_1}, $S_3$ and $S_6$ each contain three variations with different compositions of random action to examine their impact. The Subtask {\romannumeral 1} success rates (\textit{S. R.}) of these policies, along with the average pairwise $L_{1}$ distance $L_{1}^{\text{avg}}$ 
of cubes' initial position in their training datasets, are reported in \textit{Tab.}~\ref{tab:data_composition}. 
The statistical results show \textit{a strong link} between a data collection policy’s success rate and the $L_{1}^{\text{avg}}$ of its training data, with higher $L_{1}^{\text{avg}}$ generally indicating better performance.
This effect can be attributed to the fact that evenly distributed data improves performance through broader workspace coverage, with random actions reducing the distributional bias of autonomous policies (\textit{Fig.}~\ref{fig:data_compositionB}).
However, it is worth noting that random actions alone cannot always ensure close to uniform distributions (\textit{Fig.}~\ref{fig:data_compositionC}), as clustering may occur due to random sampling.
To ensure autonomous and sustainable policy bootstrapping, metrics such as $L_{1}^{\text{avg}}$ should be incorporated during episode composition, to preserve distributional balance.

\renewcommand{\arraystretch}{1.2} 
\setlength{\tabcolsep}{6pt} 
\begin{table}[h]
\centering
\label{tab:lnn}
\caption{\small Success rates and $L_{1}^{\text{avg}}$ of training datasets corresponding to the composed policies.}
\label{tab:data_composition}
\begin{tabular}{llll|lll} 
\toprule
&\textbf{$\pi_3^{\{1, 2\}}$} & \textbf{$\pi_3^{\{1, 2\}E}$} & \textbf{$\pi_3^{\{1\}}$} & \textbf{$\pi_6^{\{4, 5\}}$} & \textbf{$\pi_6^{\{4, 5\}E}$} & \textbf{$\pi_6^{\{1\}}$}\\ 
\midrule
$L_{1}^{\text{avg}}$& 0.0269 & 0.0273 & 0.0235 & 0.0422 & 0.0434& 0.0546\\
\textit{S. R.}& 0.14 & 0.18 & 0.12 & 0.27 & 0.32& 0.35\\
\bottomrule
\end{tabular}
\end{table}

\subsubsection{\textbf{The cost-effectiveness of random actions}}
We provide empirical results evaluating the cost-effectiveness of random actions using collection time efficiency (h) and collection cost (USD) for 100 successful episodes.
\textit{Tab.}~\ref{tab:cost_effectiveness_paradigm_1} shows the cost of collecting 100 successful episodes of stacking with the complete-task policies $\pi_4$ (random), $\pi_5$ (autonomous), $\pi_6^{\{4\}}$ (composed), and a proficient human demonstrator (Human) using keyboard control with the top-camera view.
The rough cost estimation is divided into three categories: robot electricity usage (Hardware, \$0.06/h), GPU inference for the data collection policy (Inference, \$0.04/h), and human labor for a demonstrator experienced in robotics (Human Labor, \$25/h). 
The estimates are based on manufacturer-reported power consumption for the UR10e robot and the NVIDIA RTX 3070 GPU, together with hourly wages for robotics operators obtained from hiring platforms including ZipRecruiter and Glassdoor.
Electricity cost is computed using the average U.S. commercial electricity rate of \$0.1747 per kWh reported by the U.S. Energy Information Administration in 2025. 
Given regional variability in labor and energy prices, these values are approximate and should be interpreted as U.S.-specific reference points.
\begin{table}[h]
        \centering
        \caption{\small Cost-effectiveness analysis of random actions.}
        \renewcommand{\arraystretch}{1.2} 
        \setlength{\tabcolsep}{3pt}{\footnotesize
        \begin{tabular}{cc|cccc}
            \toprule
            Method& Effectiveness (\textit{h}) & \multicolumn{4}{c}{Cost (\textit{\$})} \\
            &Time & Hardware & Inference & Human Labor & \textbf{Total}\\
            \midrule
            $\pi_4$ &101.8 &6.11 &4.07& 0& 10.18\\
            $\pi_5$ & 10.7 & 0.64& 0.42& 0& 1.08\\
            $\pi_6^{\{4\}}$& 5.3 & 0.32& 0.21& 0& \textbf{0.53}\\
            \midrule
            \textit{Human}& 3.8 & 0.23&0 & 95 & \textbf{95.23}\\
            \bottomrule
        \end{tabular}
        }
    \label{tab:cost_effectiveness_paradigm_1}
    \vspace{-1em}
\end{table}

Within three policy updates, the time-efficiency gap between the data collection policy bootstrapped from random actions and the human demonstrator narrows to 1.5 h.
Although controller differences may introduce discrepancies, the stacking policy bootstrapped from random actions substantially reduces costs by avoiding human labor, \textit{showing the strong promise of random actions as a cost-effective data source for the selected task.}
While these conclusions are based on a two-layer stacking task, the significant cost differences indicate the value of reassessing the real-world data collection and incorporating complementary sources for scaling, with random actions serving as one example.

\subsection{Results of Paradigm \MakeUppercase{\romannumeral 2}}
\subsubsection{\textbf{What pre-training objective works best on random exploration video frames?}}
\label{sec:objective}
\begin{figure}[h]
    \centering
    \includegraphics[width=0.8\linewidth]{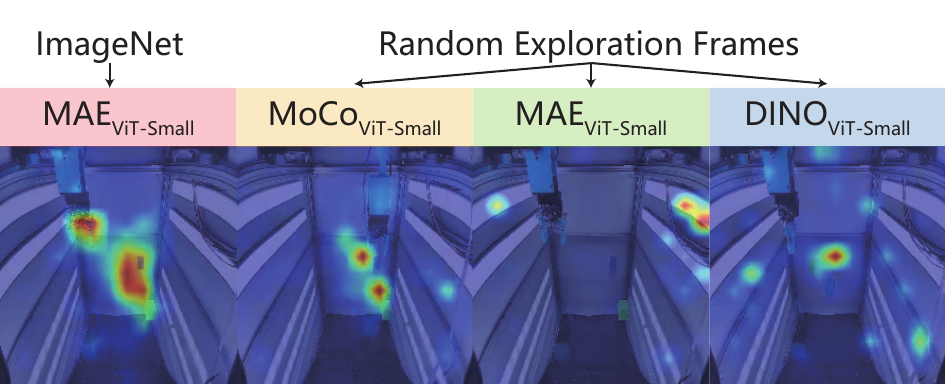}
    \caption{\small Visualization of FullGrad saliency maps for models with different pre-training objectives and datasets. Complete video sequences are available in the supplementary video.}
    \label{fig:saliency}
    \vspace{-1em}
\end{figure}

We begin by assessing the performance of pre-training on random exploration video frames under three objectives with the ViT-Small architecture: MAE (reconstruction loss), MoCo (contrastive loss), and DINO (distillation loss).
The results are summarized in \textit{Tab.}~\ref{table:architecture}, where ImageNet~\cite{deng2009imagenet} pre-trained with the three objectives is included for comparison due to its widespread use and strong results demonstrated in previous studies in self-supervised pre-training~\cite{dasari2023unbiased, paolini2014data}.
The results show that random exploration video frames pre-trained with MoCo objective outperform both MAE and DINO by \textit{a noticeable margin}. 
To further investigate this, we use FullGrad~\cite{jacobgilpytorchcam} saliency maps to visualize the impact of different objectives on visual encoders, as shown in \textit{Fig.}~\ref{fig:saliency}. 
We denote models pre-trained on ImageNet using the MoCo objective with the ViT-Small architecture as $\text{MoCo}_{\text{ViT-Small}}(\text{ImageNet})$. 
It can be seen that the activations of $\text{MAE}_{\text{ViT-Small}}(\text{ImageNet})$ and $\text{MoCo}_{\text{ViT-Small}}(\text{Random})$, the best- and second-best-performing models, concentrate mainly in the task area. 
In contrast, the activations of poor-performing $\text{MAE}_{\text{ViT-Small}}(\text{Random})$ and $\text{DINO}_{\text{ViT-Small}}(\text{Random})$ focus on the background and exhibit dispersed activations. 
This potentially suggests that: (1) MoCo’s better performance may be linked to its task-focused activations; (2) its concentrated activation likely helps with object localization, explaining good results in $S_{\text{grasping}}$ and $S_{\text{one}}$; and (3) this focus might also account for its \textit{uneven distribution} of successful episodes observed during evaluation. 
We find that $\text{MoCo}_{\text{ViT-Small}}(\text{Random})$, consistently failed to reach when the red block was in the top-right, while $\text{MoCo}_{\text{ViT-Small}}(\text{ImageNet})$ succeeded in 2-layer stacking only when the green block was in the bottom-left. 
This may be because MoCo, designed for discriminating individual frames, introduces a bias favoring lower-positioned objects, which occupy more pixels than those higher in the image.

\begin{table}[h]
\centering
\setlength{\tabcolsep}{2pt}
\caption{\small Impact of pre-training objectives on Random Exploration Video Frames (Random) and ImageNet.}
\label{table:architecture}
\begin{tabular}{lcc|cc|cc}
\toprule
Architecture & \multicolumn{2}{c|}{$\text{MAE}_{\text{ViT-Small}}$} & \multicolumn{2}{c|}{$\text{MoCo}_{\text{ViT-Small}}$} & \multicolumn{2}{c}{$\text{DINO}_{\text{ViT-Small}}$} \\ 
Dataset & Random & ImageNet & Random & ImageNet & Random & ImageNet \\ 
\midrule
$PE \downarrow$ & 0.158 & 0.048 & 0.032 & 0.037 & 0.114 & 0.253 \\ 
$S_{\text{grasping}} \uparrow$ & 0 & 0.95 & 0.8 & 0.95 & 0 & 0 \\
$S_{\text{one}} \uparrow$ & 0 & 0.95 & 0.8 & 0.95 & 0 & 0 \\
$S_{\text{two}} \uparrow$ & 0 & 0.45 & 0.25 & 0.25 & 0 & 0 \\
\bottomrule
\end{tabular}
\begin{tablenotes}
\footnotesize
\item The pre-trained weights for ImageNet are adopted from~\cite{xiao2022masked, chen2021empirical, caron2021emerging}. 
\end{tablenotes}
\vspace{-1em}
\end{table}

\subsubsection{\textbf{What architecture works best on random exploration video frames?}}
Using the MoCo objective, we conducted an additional study to identify the most effective architecture for leveraging random exploration video frames, ranging from ViT-Base~\cite{dosovitskiy2020image} to ResNet18~\cite{he2016deep}, given the considerable training cost differences between architectures.
As shown in \textit{Tab.}~\ref{table:size}, the effectiveness of MoCo on random exploration video frames is highly dependent on model capacity. 
The low performance of $\text{ViT-Base}$ compared to $\text{ResNet18}$ may result from too few training episodes relative to model size. 
Despite this, $\text{MoCo}_{\text{ViT-Small}}$ remains the most effective policy for leveraging random exploration video frames and is therefore used in the subsequent cost-effectiveness analysis.

\begin{table}[h]
\centering
\setlength{\tabcolsep}{2pt}
\caption{\small Impact of model size on MoCo objective.}
\label{table:size}
\begin{tabular}{lcccc}
\toprule
 Model & $\text{MoCo}_{\text{ResNet18}}$ & $\text{MoCo}_{\text{ResNet50}}$ & $\text{MoCo}_{\text{ViT-Small}}$ & $\text{MoCo}_{\text{ViT-Base}}$ \\ 
 Params & 16M & 35M & 43M & 108M \\
\midrule
$PE \downarrow$ & 0.033 & 0.038 & \underline{0.032} & \textbf{0.029}\\
$S_{\text{grasping}} \uparrow$ & \underline{0.6} & 0.25 &\textbf{ 0.8} & 0.15\\ 
$S_{\text{one}} \uparrow$ & \underline{0.5} & 0.15 & \textbf{0.8} & 0.05\\ 
$S_{\text{two}} \uparrow$ & \underline{0.15} & 0 & \textbf{0.25} & 0\\ 
\bottomrule
\end{tabular}
\begin{tablenotes}
\footnotesize
\item The best-performing policy is in bold, and the second-best is underscored. The parameter count for each model is calculated as the sum of its trainable parameters.
\end{tablenotes}
\vspace{-1em}
\end{table}

\subsubsection{\textbf{The cost-effectiveness of random exploration video frames}}
We analyze the cost-effectiveness of random exploration video frames by comparing $\text{MoCo}_{\text{ViT-Small}}$ with end-to-end trained policies, which are widely regarded as high performance with low training cost. 
For this comparison, we select BC-GMM~\cite{mandlekar2021matters} with a ResNet18 encoder (BC) and Diffusion Policy~\cite{chi2023diffusion} (DP), distinguished by network size and temporal reliance.
Since the collection of random exploration frames is autonomous and all policies share the same human demonstrations, our analysis centers on GPU costs and their relative performance.
Training hours are measured with NVIDIA A100 40GB GPUs, and costs are estimated at \$3/GPU hour based on typical providers (e.g., AWS, Azure). 
The results are summarized in \textit{Tab.}~\ref{table:cost_effectiveness_paradigm2}.
From an effectiveness perspective, Diffusion Policy achieves the best performance, followed by $\text{MoCo}_{\text{ViT-Small}}$, with BC performing the worst.
Diffusion Policy’s advantage may come from its use of past horizons, which supports more stable trajectory predictions and better handles the long two-layer stacking sequence.
$\text{MoCo}_{\text{ViT-Small}}$ outperforms BC, which fails largely because of inaccurate grasp point prediction.
Considering cost, the 750 hours of pre-training required by $\text{MoCo}_{\text{ViT-Small}}$ substantially undermine its cost-effectiveness. 
While pre-training can be leveraged across tasks within the same environment, transferring to a different environment requires curating 700K random exploration frames, a process less favorable than direct end-to-end training when targeting a single task in this case.

\begin{table}[h]
\centering
\setlength{\tabcolsep}{2pt}
\caption{\small Cost-effectiveness analysis of random exploration video frames.}
\label{table:cost_effectiveness_paradigm2}
\begin{tabular}{lcccc|cc}
\toprule
Model& \multicolumn{4}{c|}{Effectiveness} & \multicolumn{2}{c}{Cost (\textit{\$})} \\
 & $PE \downarrow$ & $S_{\text{grasping}} \uparrow$ & $S_{\text{one}} \uparrow$ & $S_{\text{two}} \uparrow$ & Pre-training & Policy training\\
\midrule
$\text{MoCo}_{\text{ViT-Small}}$  & 0.032 & \underline{0.8} & \underline{0.8}& \underline{0.25} & 2250 & 22.3\\
BC & \underline{0.028} & 0.5&0.5 & 0.2 & 0 & 0.6\\ 
DP & \textbf{0.020} & \textbf{0.9}&\textbf{0.9} & \textbf{0.4} & 0& 36\\ 
\bottomrule
\end{tabular}
\begin{tablenotes}
\footnotesize
\item The best-performing policy is in bold, the second-best is underscored.
\end{tablenotes}
\vspace{-1em}
\end{table}

\section{Conclusion}
\begin{figure}[h]
    \centering
    \begin{subfigure}[h]{0.2\textwidth}
    \centering
    \includegraphics[width=\textwidth]{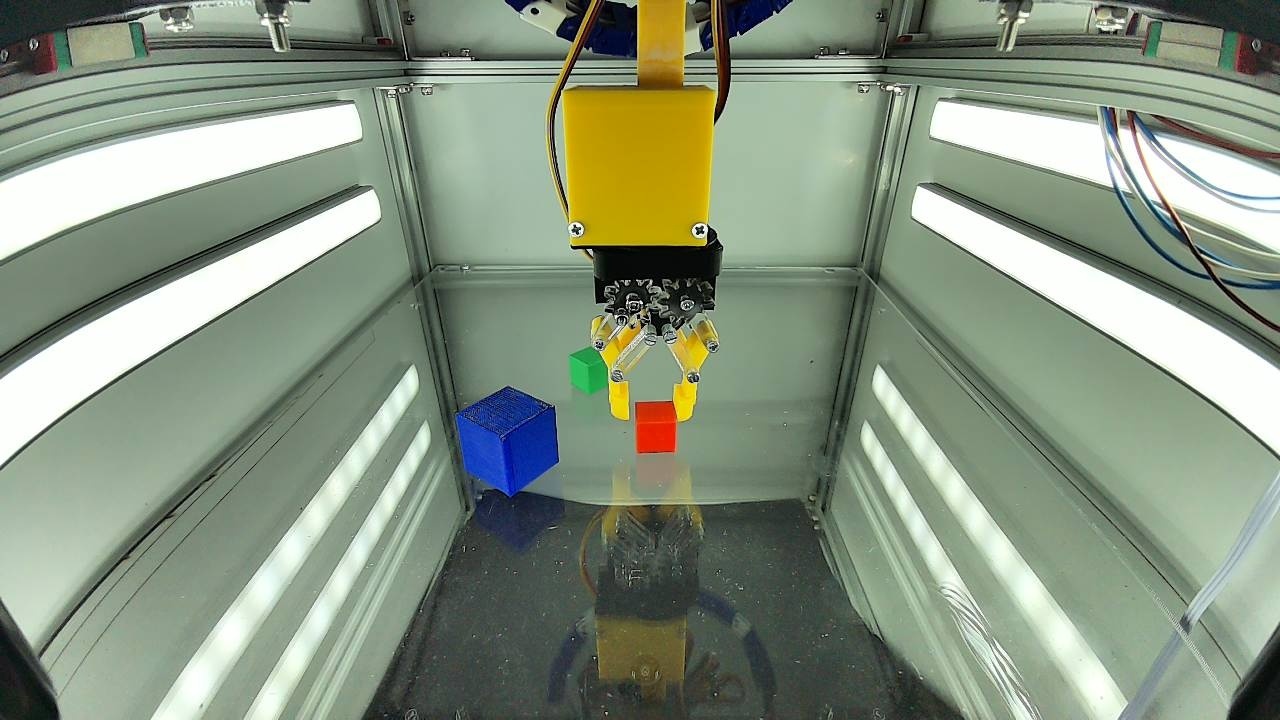}
  \end{subfigure}
  \hspace{1em}
  \begin{subfigure}[h]{0.085\textwidth}
    \centering
    \includegraphics[width=\textwidth]{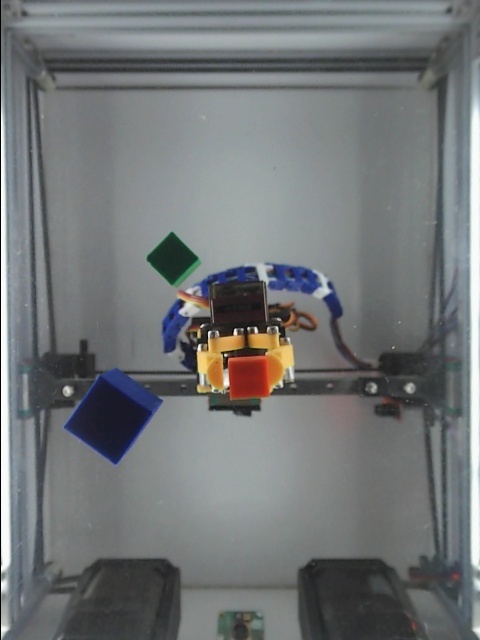} 
  \end{subfigure}
    \caption{\small Illustration of extending the current system to an assembly task: the bootstrapped policy grasps a third blue cube and stacks it at a new position through random actions.}
    \label{fig:future}
    \vspace{-1em}
\end{figure}

\textbf{Limitations and Future Work. }
This study adopts a case study approach, focusing exclusively on stacking tasks to provide large-scale real-world statistics for drawing conclusions. 
In addition, our analysis of random exploration video frames considered only individual frames, which may limit the use of dynamic information shaped by consistent environmental constraints. 
Future work will build on identified trends and strategies, such as $L_{1}^{\text{avg}}$ monitoring, to bootstrap more dexterous skills beyond grasping and stacking, extending to tasks such as assembly (\textit{Fig.}~\ref{fig:future}) and advancing from cubes to objects with irregular shapes. 
We also plan to incorporate additional sensory feedback, such as force-torque sensors and event cameras, streamed and containerized as cloud-based microservices. 
This will enable more advanced real-time monitoring, support a wider range of success patterns, and maximize the diversity benefits of random exploration data.

\textbf{Conclusion. }
In this paper, we present a large-scale experimental study using over 900 hours of autonomous robot activity to investigate the cost-effectiveness of random exploration data within the scope of imitation learning. 
Paradigm \MakeUppercase{\romannumeral 1} investigates the feasibility for autonomously bootstrapping two-layer stacking policies from random actions.
Paradigm \MakeUppercase{\romannumeral 2} evaluates the effectiveness of random exploration video frames in pre-training parameter-dense networks through three self-supervised objectives: reconstruction, contrastive, and distillation loss.
Results from 1,260 evaluations indicate that random actions are a cost-effective data source for imitation learning, and we propose strategies, including $L_{1}^{\text{avg}}$ monitoring, for managing the autonomous bootstrapping process.
Within three policy updates, the time-efficiency gap with human keyboard demonstrations is reduced to 1.5 hours per 100 successful episodes, with an estimated cost of 0.53 USD. 
Additionally, we identify the contrastive objective with a ViT-Small architecture as an effective approach for self-supervised pre-training on random exploration video frames; however, its performance is highly sensitive to initial object positions, and its cost-effectiveness is reduced by the expensive pre-training process.
We will release the $R900$ dataset, comprising 807 hours of random actions and 71 hours of random exploration video frames as an MIT-licensed open-source dataset.
The robot environment, along with the fully automated pipeline using microservices for real-time monitoring, will be remotely accessible via cloud service to support continued research.




\bibliographystyle{IEEEtran}
\balance
\bibliography{citations}

\end{document}